\documentclass[11pt]{article}
\pdfoutput=1

\usepackage{microtype}
\usepackage{graphicx}
\usepackage{subfigure}
\usepackage{booktabs} 
\usepackage{amssymb}
\usepackage{bm}
\usepackage{bbm}
\usepackage{amsmath}  
\usepackage{amsthm}
\usepackage{xcolor}
\usepackage{textcomp}
\usepackage{multirow}
\usepackage{mathtools}
\usepackage{listings}
\usepackage[margin=1in]{geometry}
\usepackage{authblk}
\usepackage[numbers]{natbib}
\usepackage{xspace}
\usepackage{comment}
\usepackage{thmtools, thm-restate}  
\usepackage{verbatim}
\usepackage{tikz}
\usetikzlibrary{arrows}
\usetikzlibrary{positioning}
\usepackage{algorithm}
\usepackage{algorithmic}
\usepackage{listings}
\usepackage{float} 
\floatstyle{ruled}
\newfloat{listing}{tbp}{lol}
\floatname{listing}{Listing}

\tikzset{
  treenode/.style = {align=center, inner sep=0pt, text centered,
    font=\sffamily},
  arn_n/.style = {treenode, circle, black, font=\sffamily\bfseries, draw=black,
    fill=white, text width=1.5em},
  arn_r/.style = {treenode, circle, black, font=\sffamily\bfseries, draw=black,
    fill=white, text width=1.0em},
  arn_x/.style = {treenode, rectangle, draw=black,
    minimum width=0.5em, minimum height=0.5em}
}

\usepackage{hyperref}

\usepackage{bm}
\usepackage{enumitem}

\newcommand*{\ShowNotes}{}

\ifdefined\ShowNotes
  \newcommand{\colornote}[3]{{\color{#1}\bf{#2 #3}\normalfont}}
\else
  \newcommand{\colornote}[3]{}
\fi

\definecolor{darkred}{rgb}{0.7,0.1,0.1}
\definecolor{darkgreen}{rgb}{0.1,0.5,0.1}
\definecolor{cyan}{rgb}{0.7,0.0,0.7}
\definecolor{dblue}{rgb}{0.2,0.2,0.8}
\definecolor{maroon}{rgb}{0.76,.13,.28}
\definecolor{burntorange}{rgb}{0.81,.33,0}
\definecolor{royalpurple}{rgb}{0.47,.31,0.66}

\ifdefined\ShowNotes
  
\else
  
\fi

\newcommand{\eg}{{\it e.g.,}\xspace}
\newcommand{\ie}{{\it i.e.,}\xspace}

\newenvironment{packeditemize}{\begin{list}{$\bullet$}{\setlength{\itemsep}{0.5pt}\addtolength{\labelwidth}{-4pt}\setlength{\leftmargin}{2ex}\setlength{\listparindent}{\parindent}\setlength{\parsep}{1pt}\setlength{\topsep}{2pt}}}{\end{list}}

\def\FWName{LowRA\xspace}

\def\nummodels{4\xspace}
\def\numdatasets{4\xspace}

\newif\ifarxiv

\title{\textbf{LowRA}: Accurate and Efficient LoRA Fine-Tuning of LLMs\\ under 2 Bits}

\author{
  Zikai Zhou$^{1}$, 
  Qizheng Zhang$^{1}$,
  Hermann Kumbong$^1$, 
  Kunle Olukotun$^{1,2}$\\
  $^1$Department of Computer Science, Stanford University\\
  $^2$Department of Electrical Engineering, Stanford University\\
  \texttt{\{zikai,qizhengz,kumboh,kunle\}@stanford.edu}
}

\begin{document}

\maketitle

\nocite{*}

\begin{abstract}

Fine-tuning large language models (LLMs) is increasingly costly as models scale to hundreds of billions of parameters, and even parameter-efficient fine-tuning (PEFT) methods like LoRA remain resource-intensive. 
We introduce LowRA, the first framework to enable LoRA fine-tuning below 2 bits per parameter with minimal performance loss. 
LowRA optimizes fine-grained quantization—mapping, threshold selection, and precision assignment—while leveraging efficient CUDA kernels for scalable deployment. 
Extensive evaluations across 4 LLMs and 4 datasets show that LowRA achieves a superior performance–precision trade-off above 2 bits and remains accurate down to 1.15 bits, reducing memory usage by up to 50\%. 
Our results highlight the potential of ultra-low-bit LoRA fine-tuning for resource-constrained environments. 

\end{abstract}


\section{Introduction}

Fine-tuning large language models (LLM) can enhance their performance on particular tasks ~\cite{wei2021finetuned,wang2022super, ziegler2019fine}, and remove unwanted behaviors like hallucinations~\cite{hu2024mitigating,liu2023mitigating} and harmful responses~\cite{bai2022training,askell2021general}.  Yet as models scale -  \eg Llama 3.1 with 405 billion parameters~\cite{dubey2024llama} and DeepSeek-V3 with 671 billion parameters~\cite{liu2024deepseek} - the cost of fine-tuning soars~\cite{hu2021lora}. 




To reduce fine-tuning costs, parameter-efficient fine-tuning (PEFT) methods \cite{zaken2021bitfit,hu2021lora,mao2021unipelt,liu2022few} freeze a model’s core weights and insert small trainable modules. We focus on LoRA \cite{hu2021lora}, which adds rank-decomposed adapters to cut computation and memory demands. Still, fine-tuning large models on a single GPU can exceed memory limits. Quantized LoRA approaches (e.g., QLoRA \cite{dettmers2024qlora}, LoftQ \cite{li2023loftq}) resolve this by quantizing the base weights with minimal loss in accuracy, enabling billions-parameter models to be fine-tuned on standard single GPUs or even mobile devices.




Despite the success of quantized LoRA in cutting memory usage, all existing works focus on LoRA fine-tuning within the range of 2 to 4 bits (per parameter), many of which are incompatible with ultra-low-bit LoRA fine-tuning~\cite{wang2024lora,meng2024pissa,dettmers2024qlora}. Further pushing down the bits (per parameter) requirement, \eg below 2 bits, has profound implications in fine-tuning and deployment in ultra-low-resource scenarios like embedded devices~\cite{shen2023agilequantactivationguidedquantizationfaster, chai2025flexquantelasticquantizationframework} and mobile phones~\cite{wang2025bitstackanysizecompressionlarge, tan2024mobilequantmobilefriendlyquantizationondevice}. However, current methods face three fundamental limitations:
\begin{itemize}
    \item \textbf{L1:} focus exclusively on coarse-grained quantization of the base model weights.
    \item \textbf{L2:} leverage quantization functions (i.e., mappings and thresholds) that assume some fixed data distribution across the entire model weight.
    \item \textbf{L3:} rely on simulated quantization, with no system support for efficient low-bit quantization. 
\end{itemize}

To unleash the full potential of quantized LoRA fine-tuning, we present \FWName{}, an accurate and efficient framework that enables 
LoRA fine-tuning down to below 2 bits (per parameter). 
\FWName{} features three major components for each of the three challenges:
\begin{itemize}
 \item \textbf{(1)} mapping/threshold function search,
 \item \textbf{(2)} fine-grained precision assignment, and 
 \item \textbf{(3)} CUDA-based kernels as quantization primitives.
\end{itemize}

Addressing \textbf{L1} and \textbf{L2} requires extra care because LoRA base weights have to work with multiple sets of adapters in real-life settings \cite{sheng2024slora,ostapenko2024towards,chen2024punica}. 
This constraint demands a powerful, task-agnostic quantization technique. 
Furthermore, optimally assigning precisions at a fine-grained granularity for LLMs calls for a scalable, low-complexity solution to handle massive parameter spaces. 
\FWName{} meets the need through a hierarchical ILP(Integer Linear Programming)-based precision assigner for performing fine-grained mixed-precision. 
Moreover, \FWName{} provides a weighted Lloyd-Max \cite{lloyd1982least,max1960quantizing}  formulation of mapping/threshold search for groupwise normalization, and achieves strong practical performance through its efficient solution.


We conduct extensive evaluation of \FWName{} across \nummodels{} choices of widely used base LLMs and \numdatasets{} choices of natural language applications, and compare \FWName against state-of-the-art baselines. 
Evaluation results demonstrate that \FWName: 
\textbf{(1)} achieves a superior performance-precision trade-off above 2 bits (per parameter) compared to baselines, and is the first method to enable accurate, efficient LoRA fine-tuning below 2 bits,
\textbf{(2)} enables substantial memory footprint reduction in fine-tuning, and
\textbf{(3)} incurs minimal additional overhead even with newly added components.

In summary, we make the following contributions:
\begin{itemize}
    \item \textbf{Identifying Key Limitations:} We identify three core limitations in existing quantized LoRA approaches, highlighting opportunities to exploit fine-grained precision assignment and mappings/thresholds.
    \item \textbf{Design and Implementation of \FWName:} We introduce \FWName, an accurate, end-to-end framework that applies fine-grained quantization to LoRA fine-tuning of LLMs. 
    We detail its key design choices—including a mappings/thresholds learner, precision assigner, and practical programming primitives—ensuring both efficiency and high performance.
    \item \textbf{Better Performance-Precision Trade-Off:} Comprehensive evaluations show that \FWName outperforms baselines in performance-precision trade-off, enabling an average 0.86-bit reduction per parameter without sacrificing performance.
    \item \textbf{Memory Footprint Reduction:} \FWName cuts memory usage by 30–50\% during fine-tuning and deployment with minimal performance loss.
    Moreover, it enables fine-tuning and deploying LLMs in ultra-resource-constrained settings at as low as 1.15 bits. 
    \item \textbf{Open-Source Framework:} We will open-source our framework and artifacts to spur further research in ultra-low-bit LoRA fine-tuning.
\end{itemize}

This paper is organized as follows: Section~\ref{sec:background} introduces LoRA fine-tuning, quantization for LoRA, and three key limitations of existing quantized LoRA methods. Section~\ref{sec:e2eframework} presents the \FWName{} end-to-end workflow, while Section~\ref{sec:design} discusses its key design insights and benefits. Sections~\ref{sec:learner} and \ref{sec:assigner} detail the mapping/threshold learner and the fine-grained precision assignment, respectively. Finally, we describe our evaluation setup, results, and takeaways in Section~\ref{sec:eval}.
\section{Background and Motivation}
\label{sec:background}

\begin{figure*}[ht]
\begin{center}
\centerline{\includegraphics[width=\columnwidth]{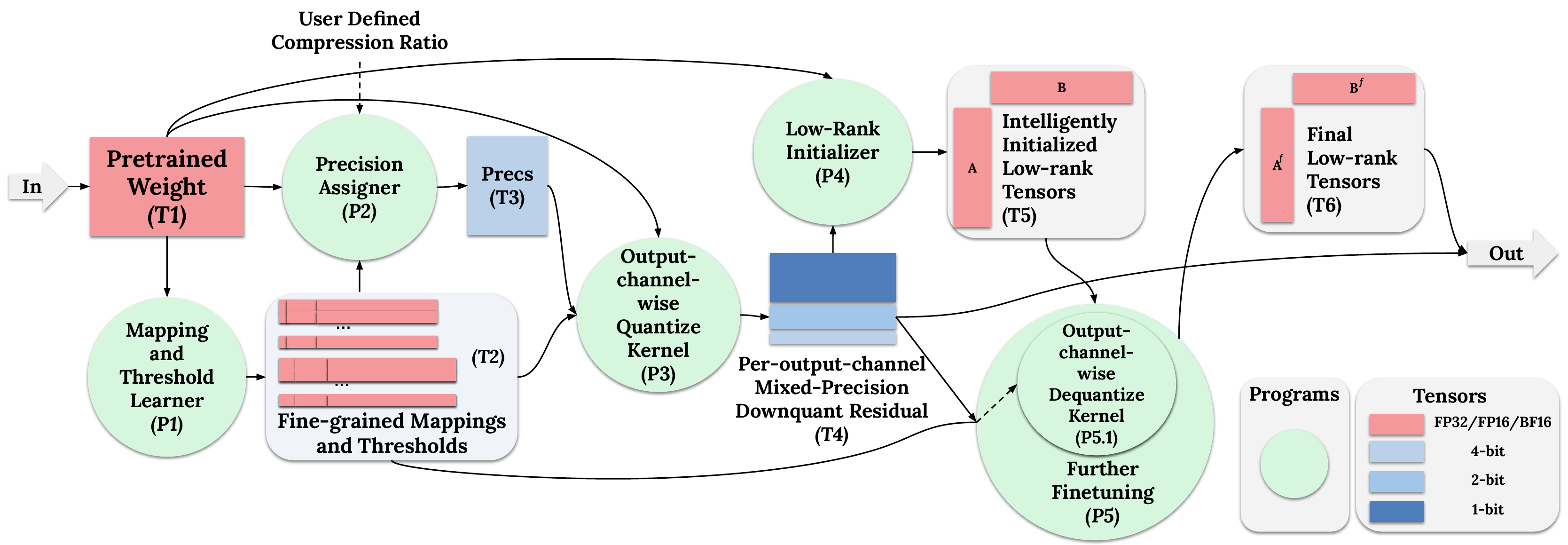}}
\vspace{-10pt}
\caption{End-to-end workflow of \FWName{}.}
\label{fig:e2eworkflow}
\end{center}
\vspace{-30pt}
\end{figure*}

\subsection{Low-Rank Adaptation (LoRA) of LLMs}

Fine-tuning large language models (LLMs) allows us to adapt pre-trained LLMs to particular tasks or domains~\cite{wei2021finetuned,wang2022super, ziegler2019fine}.
This process usually requires changing all model parameters, which can be prohibitively expensive (in terms of compute and memory) when the number of model parameters increases.
Low-Rank Adaptation (LoRA) \cite{hu2021lora} tackles this by freezing the base weights and introducing a small set of trainable ``adapter'' parameters, drastically reducing memory and compute requirements for fine-tuning.


\subsection{Quantization for LoRA Fine-Tuning}

Quantized LoRA fine-tuning further cuts memory usage by quantizing the base model weights without hurting performance. QLoRA \cite{dettmers2024qlora} introduces a \textit{NormalFloat} format to backpropagate through a 4-bit quantized backbone, while LoftQ \cite{li2023loftq} and ApiQ \cite{liao2024apiq} jointly optimize quantized base weights and adapter initializations under a unified objective. These advances unlock fine-tuning and deployment of LLMs on low-resource platforms like embedded devices \cite{shen2023agilequantactivationguidedquantizationfaster, chai2025flexquantelasticquantizationframework} and mobile phones \cite{wang2025bitstackanysizecompressionlarge, tan2024mobilequantmobilefriendlyquantizationondevice}.


\subsection{Limitations of Existing Quantized LoRA Methods} \label{sec:limitations}
Despite the early promise of recent work in quantized LoRA like QLoRA and LoftQ, we observe three major limitations that prevent us from fully realizing the potential of memory-efficient LLM fine-tuning. 

\paragraph{L1: Coarse-Grained Precision Assignment.}
Existing approaches typically apply a single quantization precision to an entire weight matrix or multiple layers. For instance, QLoRA uses uniform 4-bit quantization across all base weights, while LoftQ adopts a layerwise mixed-precision scheme (e.g., higher precision for earlier layers, lower for later layers). Our findings (\S\ref{sec:assigner}) suggest that truly unlocking ultra-low-bit fine-tuning requires a finer-grained assignment strategy, potentially at the sub-layer or sub-matrix level.

\begin{figure}[ht]
\begin{center}
\centerline{\includegraphics[width=0.8\columnwidth]{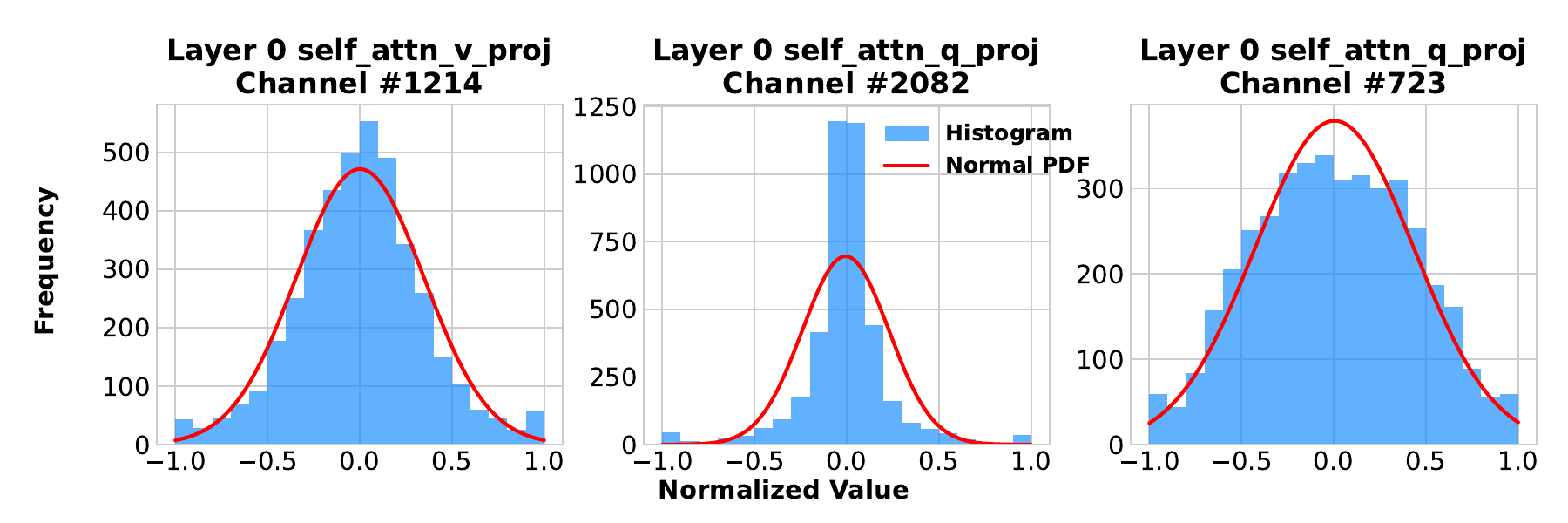}}
\vskip -0.2in
\caption{Distributions of normalized parameters in different output channels sampled from the first layer of Llama2-7b.}
\label{fig:channel_distribution}
\end{center}
\vspace{-20pt}
\end{figure}

\paragraph{L2: Discrepancy in Data Distribution.}
Most quantized LoRA methods use a globally shared data format—such as QLoRA’s \textit{NormalFloat}, which assumes a roughly normal distribution. However, Figure~\ref{fig:channel_distribution} reveals that groupwise normalization at a per-channel level often deviates significantly from a global normal distribution. To preserve accuracy, more localized quantization/dequantization approaches are needed.

\paragraph{L3: Lack of High-Performance Quantization Primitives.} \label{sec:primlimit}

Most quantized LoRA methods and related quantization studies \cite{li2023loftq, qin2024accurate,shen2020q,bai2020binarybert} rely on \textit{simulated} quantization \footnote{using floating-point values to mimic discrete quantization levels throughout training or fine-tuning.}, lacking native hardware support for sub-4-bit or flexible mixed-precision operations. For instance, LoftQ requires eight A100 GPUs even for smaller LLMs. Such reliance on simulation inflates resource requirements and impedes practical deployment (see Appendix \ref{app:githubissues} for related Github issues), as no existing system offers efficient low-bit or adaptive-precision kernels tailored to LoRA.

\section{The \FWName{} Framework}\label{sec:e2eframework}

In this section, we provide an overview of the \FWName{} end-to-end workflow, illustrated in Figure~\ref{fig:e2eworkflow}. The process begins with the pretrained model weights \textbf{\textit{(T1)}}. We feed each layer of these weights into a dedicated mapping and thresholds learner \textbf{\textit{(P1)}}, which produces optimized per-output-channel mappings and thresholds, denoted \textbf{\textit{(T2)}}.
These mappings and thresholds, along with the pretrained weights, are then processed by a two-step ILP quantizer \textbf{\textit{(P2)}} to determine the optimal precision assignments \textbf{\textit{(T3)}} for each output channel.

Next, the output-channel-wise quantize kernel \textbf{\textit{(P3)}}, which supports custom quantization thresholds, uses the derived thresholds \textbf{\textit{(T2)}} and the assigned precision levels \textbf{\textit{(T3)}} to quantize the weights. We calculate the quantization errors arising from this step and apply low-rank tensor initialization \textbf{\textit{(P4)}}. Techniques for intelligent low-rank initialization include LoftQ  \cite{li2023loftq} and PiSSa \cite{meng2024pissa} (Appendix \ref{app:lowrankinit}) which generates low-rank tensors \textbf{\textit{(T5)}} designed to absorb quantization errors during initialization. In our implementation, we opt for LoftQ \cite{li2023loftq} as experiments show that they give better performance in the low-bit range.

The resulting mixed-precision quantized weights \textbf{\textit{(T4)}} and the initialized low-rank tensors \textbf{\textit{(T5)}} are then passed to a fine-tuning module \textbf{\textit{(P5)}}, which relies on an output-channel-wise dequantize kernel \textbf{\textit{(P5.1)}} to recover the base weights from their quantized form. Following the approach used in LoRA and QLoRA, the base weights \textbf{\textit{(T4)}} remain frozen, and only the low-rank tensors \textbf{\textit{(T5)}} are trained. Finally, we obtain the updated low-rank tensors \textbf{\textit{(T6)}}, which are output together with the quantized base weights (and associated quantization state) \textbf{\textit{(T4)}}.



\section{Discussion about Design Choices} \label{sec:design}


In this section, we discuss various design choices in the \FWName framework, as well as system and hardware support. 

\subsection{Insights behind \FWName{} Design Choices} \label{sec:designchoices}

\paragraph{Per-Output-Channel Quantization} 
In LLMs, linear layers often exhibit substantially more variation across output channels than across input channels. For instance, in Llama2-7b \cite{touvron2023llama}, the average standard deviation along the output channel is 2.20× higher than along the input channel (see Appendix~\ref{app:channel_var}). As a result, grouping parameters by output channel and assigning a unique precision to each group—i.e., per-output-channel quantization—more effectively captures their diverse distributions.

\paragraph{Groupwise Normalization} 
Each output channel may still exhibit significant internal variability even with per-output-channel quantization. 
To address this, groupwise normalization is often used to allow each group of elements share a separate scale. 
We follow QLoRA's design of using 64-element normalization scaled by the absmax (i.e., maximum absolute value) in each group \cite{dettmers2024qlora}. \label{sec:groupnorm}

\paragraph{Data-Free Post-Training Quantization} Unlike quantization-aware training (QAT) \cite{esser2019learned,yang2021bsq,jeon2024l4q,savarese2022not}, our approach adds no overhead to fine-tuning. By automatically searching for quantization mappings and thresholds, it frees users from manual tuning \cite{savarese2022not,zhou2023sysmol}, saving both development and computation resources. Moreover, contrary to some methods that vary compression ratios over time \cite{savarese2022not,yang2021bsq}, \FWName maintains a consistent compression ratio, ensuring persistent memory savings during fine-tuning. 

\begin{figure}[ht]
\begin{center}
\centerline{\includegraphics[width=0.7\columnwidth]{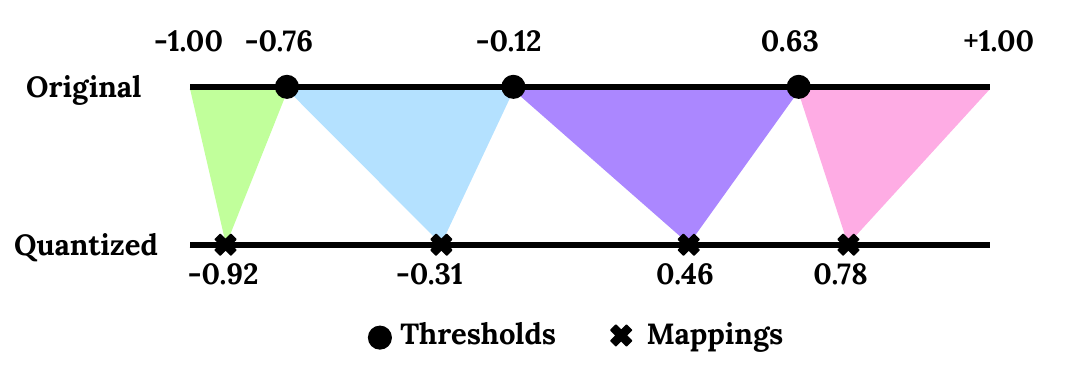}}
\vspace{-10pt}
\caption{Roles of \textbf{mappings} and \textbf{thresholds} in quantization. Circles represent thresholds whereas crosses represent mappings. Colored Triangles represent the process of converting a range of original/unquantized real values - partitioned by thresholds - to the mapped values corresponding to each quantization level. 
}
\label{fig:mappingthreshold}
\end{center}
\vspace{-20pt}
\end{figure}

\paragraph{Per-Output-Channel Thresholds and Mappings} 
Figure \ref{fig:mappingthreshold} visualizes the roles of thresholds and mappings in the process of quantization. Thresholds refer to the boundary points (``bin edges") that partition the continuous domain of normalized parameters into discrete intervals and thus specific bitstring encodings. 
Mappings, on the other hand, specify the representative values assigned to each encoded bitstring and thus the intervals. As discussed in \S\ref{sec:limitations}, fine-grained designs of quantization mappings and thresholds could lead to significantly more accurate approximation and reconstruction of parameters. 
\FWName{} allows each output channel to adopt a different combination of mappings and thresholds for more precise fine-grained quantization.

\paragraph{Data-Free One-Shot Post-Training Quantization}
Most quantization-aware training (QAT) techniques achieve higher task performance
by incurring additional training overhead and learning task-specific quantization 
parameters~\cite{esser2019learned,yang2021bsq,jeon2024l4q,savarese2022not}.
Similarly, many post-training quantization methods require a calibration set 
for quantization scheme learning~\cite{liao2024apiq,hubara2021accurate}.
In contrast, \FWName{} uses \emph{data-free one-shot post-training quantization}, 
enabling reusable quantization schemes and quantized parameters, minimal hyperparameter 
tuning, and negligible fine-tuning overhead.
This design is particularly suited to LoRA fine-tuning because: (1) Task-dependent learning is confined to the adapters, (2) LoRA base weights are often shared across multiple adapters, and (3) LoRA primarily targets resource-constrained fine-tuning scenarios. 


\paragraph{User-Defined Compression Ratios} 
Quantized LoRA methods see heavy use in tight resource settings - \eg limited-memory GPUs or on-device scenarios - where specifying a precise compression ratio is pivotal. By tailoring each parameter’s bit precision, thresholds, and mappings, \FWName directly aligns compression with real-world resource budgets, ensuring feasibility and efficiency even under strict constraints. Furthermore, because \FWName fixes the ratio in a single pass, it obviates the extensive hyper-parameter tuning needed by alternative methods to find acceptable compression–accuracy trade-offs \cite{savarese2022not,zhou2023sysmol}.

\paragraph{Using LoftQ as Low-Rank Intializer}
Researchers have found that the initialization of low-rank tensors is crucial to the effectiveness of LoRA fine-tuning, especially when it comes to ultra-low-bit quantized base weight \cite{li2023loftq, meng2024pissa,wang2024lora}. LoftQ \cite{li2023loftq} and PiSSA \cite{meng2024pissa} are two notable initialization techniques for quantized LoRA (see Appendix \ref{app:lowrankinit} for a detailed introduction).  While PiSSA  purports faster convergence than LoftQ, our experiments consistently show LoftQ outperforming PiSSA. As illustrated by the sample data points in Table \ref{tab:init_compare}, PiSSA fails to achieve reasonable task performance at lower bit ranges. This aligns with our intuition that performing quantization rather than SVD first allows the low-rank tensors to better absorb quantization errors. Our findings also corroborate points raised in the LoftQ appendix. Since our main objective is to enable lower-precision fine-tuning and deployment, we opt to use LoftQ as our low-rank initializer. Following the recommendation from LoftQ, we use five alternating steps for initialization.

\begin{table}[ht]
    \centering
    \resizebox{0.6\columnwidth}{!}{%
    \begin{tabular}{llcccc}
        \toprule
        \multicolumn{2}{l}{\textbf{Setup}} 
        & \multicolumn{2}{c}{\textbf{Llama-7b}} 
        & \multicolumn{2}{c}{\textbf{Llama-13b}}\\
        \cmidrule(lr){1-2} \cmidrule(lr){3-4} \cmidrule(lr){5-6}
        \textbf{Method} & \textbf{Dataset} & \textbf{2-bit} & \textbf{4-bit} & \textbf{2-bit} & \textbf{4-bit} \\
        \midrule
        PiSSA & \multirow{2}{*}{WikiText-2 (↓)} & 1919.63 & 5.53 & 1825.68 & 5.05 \\
        LoftQ &                                & 8.63 & 5.26 & 7.27 & 4.79 \\
        \bottomrule
    \end{tabular}%
    }
    \caption{Perplexities of PiSSA and LoftQ as initialization methods on WikiText-2. 
    Quantization is performed at 2 bits or 4 bits per parameter. Lower values indicate better performance ($\downarrow$).}
    \label{tab:init_compare}
\end{table}


\paragraph{Adapting \FWName{} to Production Use Cases} 
Many production use cases, \eg batched inference in data centers, require fixed quantization mappings \cite{li2024svdqunat,zhao2024atom}. 
\FWName{} can be adapted to such scenarios by keeping only the thresholds learnable, which is shown to be useful in enhancing model performance \cite{liu2022nonuniform}. 
To maximize performance with task-agnostic reusable base weight, \FWName{} can be extended to use the same set of thresholds for multiple adapters but learn mappings for each downstream task. 
In other words, each adapter can be connected to the base weight together with a dedicated base weight decoding mapping for that downstream task. 
Nevertheless, this would demand higher development and computation costs. 
In our implementation and experiments, we adopt the same set of learned thresholds and learned mappings for a single base weight for the proof of concept.

\subsection{System and Hardware Support}

Building on the limitations noted in \S\ref{sec:primlimit}, we implement practical CUDA-based primitives that support both low-bit and mixed-precision LoRA fine-tuning with maximum flexibility (details in Appendix~\ref{app:systemsupport}). Notably, the added kernel generalization incurs only negligible overhead in end-to-end inference, as quantization/dequantization constitutes a minimal portion of the total compute cost.
\section{Mapping and Threshold Learner} \label{sec:learner}


In this section, we introduce the mapping and threshold learner in \FWName. Because we want the final base weights to remain task-agnostic and thus reusable across multiple adapters, we adopt a simple approach that minimizes the \emph{mean squared error}\footnote{We define \textbf{SE} (\textbf{S}quared \textbf{E}rror) as 
  $(\hat{y} - y)^2$, i.e., the squared difference between the predicted 
  and ground-truth values. Therefore, \textbf{MSE} is the mean of these 
  squared errors over the dataset.} (\emph{MSE}) in each output channel. As discussed in \S\ref{sec:designchoices}, one could learn separate decoding mappings for each downstream task (or adapter set), but at a higher fine-tuning cost. We therefore propose an efficient design for the mapping/threshold learner that avoids this expense.

\paragraph{Weighted Lloyd-Max Algorithm.}

We cast the problem of searching for the optimal quantization mappings and thresholds for each output channel to minimize MSE as a Weighted Lloyd-Max Problem. 
A detailed description of this algorithm can be found in Appendix \ref{sec:lloyd-max}.

\paragraph{Weighted Lloyd's for LoRA Quantization.}
As discussed in \ref{sec:designchoices}, we perform groupwise normalization to give more scale to quantization within each output channel. To recap, with groupwise normalization (Section \ref{sec:groupnorm}), each block of weights is scaled by the block-wise maximum absolute value ($absmax$).  Specifically, if we denote the set of original weights in a block by \(\{ x_i \}\)  and its maximum absolute value by $absmax$, then we treat $absmax$ as a per-block weight in the Weighted Lloyd-Max algorithm. By assigning them proportionally larger weights, the algorithm ``pays more attention'' to those blocks and adjusts their bin thresholds and centroids accordingly. Consequently, blocks whose values have smaller magnitudes (and thus smaller $absmax$) are penalized less, striking a balance across all blocks to minimize the overall quantization error in QLoRA.

In our application of the Weighted Lloyd's algorithm to LoRA base weight quantization, we initialize the thresholds as those used by \textit{NormalFloats} \cite{dettmers2024qlora} for 2-bit and 4-bit precisions and use $0.0$ as the initial threshold for 1-bit quantization\footnote{Separately defined as NormalFloats lack a 1-bit representation}\footnote{See the initial values in Appendix \ref{app:lloydinit} }. Then, at each iteration, we recompute the quantization mappings as the weighted centroids of the assigned data and recompute the thresholds as the midpoints between consecutive mapping (centroid) values\footnote{In our implementation, we set number of iterations to 2}. We output the last-computed quantization mappings and thresholds when max iteration is reached or the MSE stops going down.





In our current implementation, we take the average of all thresholds to preserve distribution and prevent instability in the interaction with the Low-Rank Initializer.
\section{Mixed-Precision Quantization: Channelwise Precision Assignment} \label{sec:assigner}

In this section, we present how mixed-precision quantization assignment is conducted in \FWName. In light of the aforementioned task-agnostic requirement, a simple yet effective objective for defining this problem is the minimization of the overall \textit{Summed Square Error}\footnote{We define \textbf{SSE} (\textbf{S}ummed 
\textbf{S}quare \textbf{E}rror) as the sum of the squared differences between 
the original values and their quantized counterparts, i.e., 
$\sum_{i} \bigl(x_i - \widehat{x}_i\bigr)^2$.} (\textit{SSE}) considering the regular structure of transformer-based architectures \cite{waswani2017attention,dubey2024llama,touvron2023llama}. Such formulation can serve as an effective proxy to retain more information for the harder-to-quantize channels in weights.

\begin{figure}[ht]
\begin{center}
\centerline{\includegraphics[width=0.7\columnwidth]{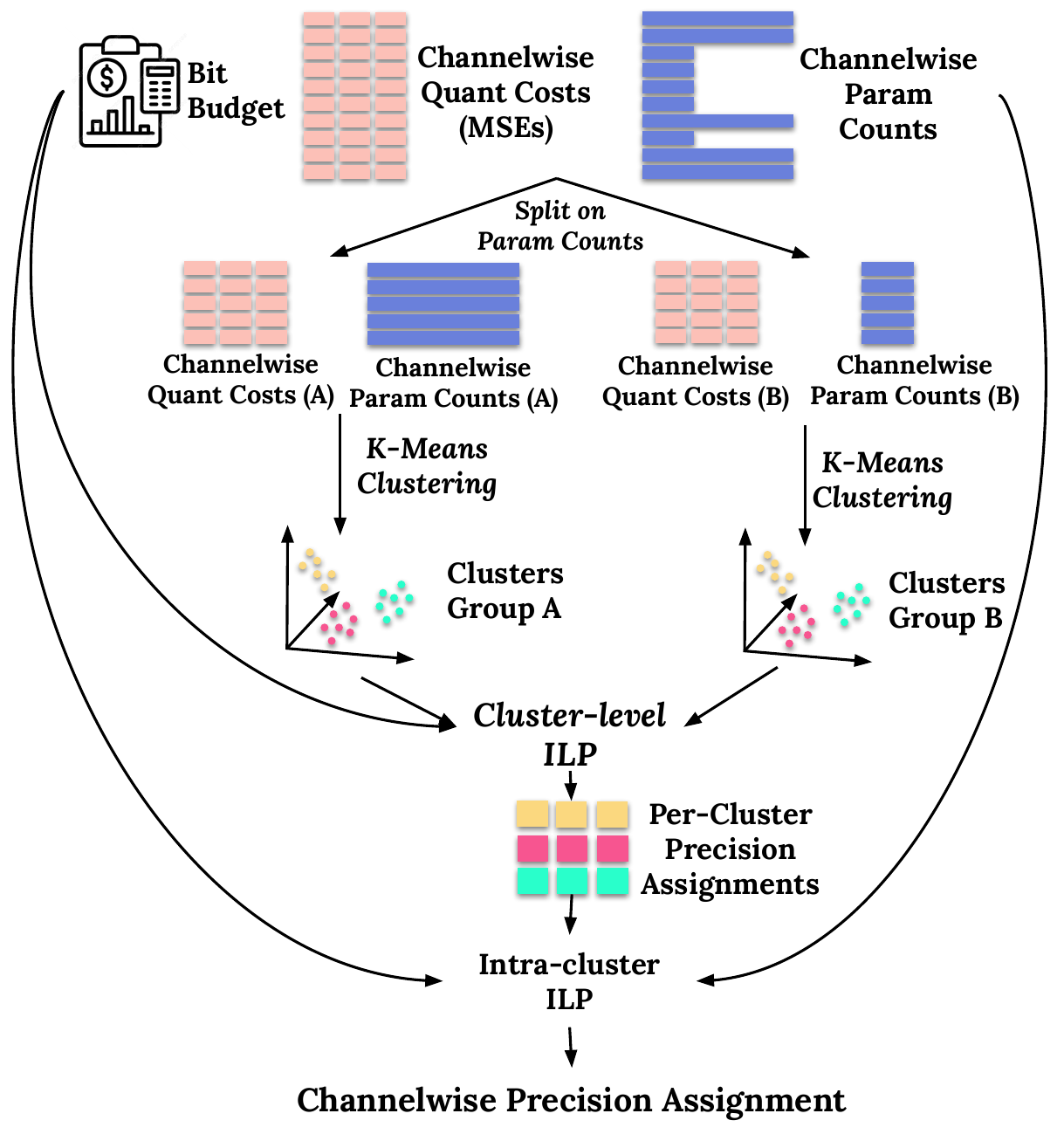}}
\caption{Two-step ILP-based Workflow for Channelwise Precision Assignment}
\label{fig:twostepilp}
\end{center}
\end{figure}

One can observe that finding the optimal mixed-precision scheme (w.r.t. SSE) can be formulated as an ILP. However, due to the large number of output channels in LLMs, a direct solver-based approach becomes prohibitively expensive. For instance, solving more than five layers of LLaMA-2-7B fails to finish within ten hours. To address this limitation, we propose a two-level ILP workflow (Figure~\ref{fig:twostepilp}, Algorithm~\ref{alg:twostepilp}) that retains the benefits of ILP-based methods while ensuring reasonable complexity.


\begin{algorithm}[ht]
\setlength{\abovecaptionskip}{-8pt}
\setlength{\belowcaptionskip}{-10pt}
\caption{Channelwise Precision Assignment}
\begin{algorithmic}[1]
  \STATE \textbf{Input}: $N$, distinct $w^{(1)},\dots,w^{(K)}$, partition $\{I_k\}$, MSE$(i,p)$, total budget $B_{\text{total}}$
  \STATE \textbf{Step 1.} Compute $W_k = \sum_{i\in I_k} w_i$, then $W_{\text{sum}} = \sum_{k=1}^K W_k$
  \STATE \textbf{Step 2.} $B_k \leftarrow B_{\text{total}} \times \frac{W_k}{W_{\text{sum}}}$ for $k=1,\dots,K$
  \FOR{$k = 1$ to $K$}
    \STATE \textbf{Step 3.} Cluster channels in $I_k$ (e.g.\ K-Means on MSE features) into $K_k$ clusters
    \STATE \textbf{Step 4.} \emph{Cluster-Level ILP}: decide how many channels in each cluster get each bitwidth, subject to $B_k$
    \STATE \textbf{Step 5.} \emph{Intra-Cluster ILP}: within each cluster, assign specific channels to bitwidths
  \ENDFOR
  \STATE \textbf{Step 6.} Combine final bitwidths into $b_1, \ldots, b_N$
  \STATE \textbf{Output:} $(b_1,\dots,b_N)$, total SSE, actual bits used
\end{algorithmic}
\label{alg:twostepilp}

\end{algorithm}

\subsection{Preprocessing for the Pipeline (Step 1-3)}

To preprocess channels for the hierarchical ILP pipeline, we first compute each channel’s MSE under 1-, 2-, and 4-bit quantization. Next, we split channels by parameter count, which in LLMs typically yields two distinct sizes (\eg 4096 and 11008 for LLaMA-2-7B). Within each group, we then apply three-dimensional K-Means clustering\footnote{We use 300 as the maximum number of iterations} (based on the three computed MSE values), forming 128 clusters per group in our implementation.

\subsection{Cluster-Level ILP (Step 4)}
Formed clusters first go through the following cluster-level ILP to be assigned budgets of 1-bit, 2-bit, and 4-bit channels.

Consider \(C\) clusters, each with \(S_c\) channels \(\bigl(c = 1,\dots,C\bigr)\). Let \(\mathcal{P} = \{1,2,4\}\) be the available bit-precisions\footnote{For LLaMA, restricting to 2 and 4 bits outperformed including 1 bit for bpp $\geq$2.0, so we adopt this configuration.}. For each cluster \(c\) and precision \(p \in \mathcal{P}\), define:
\begin{packeditemize}
    \item \(\text{cost}_{c,p}\): mean quantization error (e.g., mean-squared error) \emph{per channel} in cluster \(c\) if all channels in that cluster are assigned to precision \(p\), scaled by the number of weight parameters per channel in that cluster \footnote{Contrary to LoftQ’s suggestion, per-layer cost weighting based on layer index proved suboptimal in our experiments.}.
    \item \(y_{c,p} \in \mathbb{Z}_{\ge 0}\): decision variable representing the number of channels in cluster \(c\) that will be assigned precision \(p\).
\end{packeditemize}

We define a global bit-budget \(B\) (i.e., total permissible bits across all clusters). 
Let \(\beta(p)\) be the bit-precision value (e.g., \(\beta(1)=1\), \(\beta(2)=2\), \(\beta(4)=4\)). 
To enforce the bit budget, we multiply \(\beta(p)\) by the channel parameter count \(\omega_c\) for cluster \(c\), and then by the number of channels \(y_{c,p}\). As we split channels based on the number of parameters within each channel, each channel in a cluster \(c\) shares the same \(\omega_c\). 

\begingroup
\[
\begin{aligned}
  &\text{Minimize} 
  && \sum_{c=1}^C \sum_{p \in \mathcal{P}} \bigl(\text{cost}_{c,p}\bigr)\,y_{c,p} \\
  &\text{subject to} 
  && \sum_{p \in \mathcal{P}} y_{c,p} = S_c, \quad c = 1,\dots,C,\\
  &&& \sum_{c=1}^C \sum_{p \in \mathcal{P}} \bigl(\beta(p)\,\omega_c\bigr)\;y_{c,p} 
      \le B,\\
  &&& y_{c,p} \in \mathbb{Z}_{\ge 0}, \quad 0 \le y_{c,p} \le S_c.
\end{aligned}
\]
\endgroup

This formulation seeks to minimize the total weighted quantization error by choosing, for each cluster \(c\), how many of its channels \(y_{c,p}\) are assigned to each precision level \(p\). The constraints ensure that every channel of a cluster is allocated exactly once, the total bits used do not exceed the overall budget \(B\), and that the decision variables remain non-negative integers and do not exceed the number of channels in their respective clusters.

\subsection{Intra-Cluster ILP (Step 5)}
Once the cluster-level ILP decides how many channels \(\{y_{c,p}\}\) in each cluster \(c\) should be assigned to each bit precision \(p\), a second ILP distributes these assignments to each channel within each cluster.

Let \(S_c\) denote the total number of channels in cluster \(c\). For channel \(i \in \{1,\dots,S_c\}\) in cluster \(c\), we define the precomputed mean-squared error at precision \(p\) as
\(\text{MSE}(i,p) = \) (precomputed quantization  error of channel \(i\) at precision \(p\)).
We define binary decision variables \( x_{i,p} = 1 \) if channel \( i \) is assigned bit precision \( p \); otherwise, \( x_{i,p} = 0.\)

\begingroup
\[
\begin{aligned}
&\text{Minimize} && 
\sum_{i=1}^{S_c} \sum_{p \in \mathcal{P}} \text{MSE}(i,p)\, x_{i,p}
\\
&\text{subject to} &&
\sum_{p \in \mathcal{P}} x_{i,p} \;=\; 1 
\quad \forall\, i \in \{1,\dots,S_c\},
\\
&&& \sum_{i=1}^{S_c} x_{i,p} \;=\; y_{c,p}
\quad \forall\, p \in \mathcal{P},
\\
&&& x_{i,p} \in \{0,1\}
\quad \forall\, i \in \{1,\dots,S_c\},\; p \in \mathcal{P}.
\end{aligned}
\]
\endgroup

This formulation constitutes the intra-cluster ILP. The objective minimizes the total quantization error, where \(\text{MSE}(i,p)\) is the precomputed mean-squared error for channel \(i\) at bit precision \(p\). The first constraint ensures that each channel is assigned to exactly one precision. The second constraint enforces that the number of channels assigned to each precision \(p\) matches the counts \(y_{c,p}\) determined by the cluster-level ILP. Finally, the binary constraint stipulates that each decision variable \(x_{i,p}\) is either 0 or 1.

\noindent
This intra-cluster ILP enforces that the required number of channels (from the cluster-level ILP) is assigned to each bit precision and minimizes local MSE within the cluster.

\textbf{Efficiently Leveraging ILP Solvers.} In Step 6, we collect the assigned per-channel precisions. By employing this two-step hierarchical approach, we capitalize on the strengths of ILP solvers while maintaining minimal computational overhead.
\section{Evaluation} \label{sec:eval}

We evaluate \FWName across four datasets spanning natural language generation, multi-turn conversation, and long-context text summarization, demonstrating that:
\begin{packeditemize}
    \item 
    \textbf{Better performance at the same precision:} \FWName outperforms all baselines below 4-bit and matches their performance at 4-bit (\S\ref{subsec:key_results}).
    \item 
    \textbf{Same performance at lower precision:} \FWName achieves comparable performance while reducing precision by 0.86 bits per parameter on average (\S\ref{subsec:key_results}).
    \item 
    \textbf{First method to fine-tune LoRA under 2 bits:} \FWName enables fine-tuning down to 1.75 bits on LLaMA-2-7B, LLaMA-2-13B, and BART-large, and 1.15 bits on LLaMA-30B (\S\ref{subsec:ultra-low-bit}).
    \item 
    \textbf{Substantial memory savings:} \FWName reduces memory usage by 30–50\% in fine-tuning and deployment, with minimal performance loss compared to QLoRA (\S\ref{subsec:memory-imp}).
    \item 
    \textbf{Minimal overhead:} The additional one-time preprocessing costs in \FWName are negligible (Appendix \ref{app:overhead}).
\end{packeditemize}

\begin{table*}[ht]
    \centering
    \resizebox{\textwidth}{!}{%
    \begin{tabular}{llcccccccc}
        \toprule
        \multirow{2}{*}{Method} & \multirow{2}{*}{Bit} & \multicolumn{2}{c}{LLaMA-2-7B} & \multicolumn{2}{c}{LLaMA-2-13B} & \multicolumn{2}{c}{BART-large} \\
        \cmidrule(lr){3-4} \cmidrule(lr){5-6} \cmidrule(lr){7-8}
        & & WikiText-2 & OASST1 & WikiText-2 & OASST1 & XSUM & CNN/DailyMail \\

        \cmidrule(lr){3-5} \cmidrule(lr){6-8}\cmidrule(lr){9-10}
        & & ppl.$\downarrow$/acc.$\uparrow$ & ppl.$\downarrow$ & ppl.$\downarrow$/acc.$\uparrow$ & ppl.$\downarrow$ & \multicolumn{2}{c}{ROUGE1$\uparrow$/ROUGE2$\uparrow$/ROUGEL$\uparrow$}  \\
        \midrule
        QLoRA & 4.00 & 6.22 / 0.583 & 3.52 & \textbf{4.79} / \textbf{0.628} & 3.25 & 39.07 / 16.31 / 31.09 & \textbf{41.18 / 18.32 / 27.58} \\
        LoftQ & 4.00 & 5.26 / \textbf{0.613} & \textbf{3.48} & \textbf{4.79} / \textbf{0.628} & 3.23 & \textbf{40.34} / 17.06 / 31.92 &  41.12 / 18.29 / 27.54\\
        \textbf{\FWName} & 4.00 & \textbf{5.25} / 0.612 & \textbf{3.48} & \textbf{4.79} / \textbf{0.628} & \textbf{3.23} & 40.27 / \textbf{17.18} / \textbf{32.06} & 40.95 / 18.12 / 27.54 \\
        \midrule
        QLoRA & 3.00 & 7.13 / 0.566 & 4.56 & 6.06 / 0.588 & 3.88 & 17.60 / 2.68 / 13.93 & 15.34 / 1.12 / 10.44 \\
        LoftQ & 3.00 & 6.87 / 0.571 & 4.42 & 5.91 / 0.591 & 3.79 & 37.23 / 14.34 / 29.30 & 40.47 / 17.75 / 26.88 \\
        \textbf{\FWName} & 3.00 & \textbf{5.84} / \textbf{0.593} & \textbf{3.87} &  \textbf{5.24} / \textbf{0.611} & \textbf{3.50} &\textbf{38.84} / \textbf{15.68} / \textbf{30.57} & \textbf{40.85 / 18.12 / 27.23}\\
        \midrule
        QLoRA & 2.50 & 8.05 / 0.546 & 5.17 & 6.84 / 0.568 & 4.36 & 15.33 / 1.97 / 12.55 & 13.68 / 1.04 / 9.99 \\
        LoftQ & 2.50 & 7.72 / 0.552 & 4.98 & 6.70 / 0.572 & 4.21  & 34.48 / 12.26 / 27.05 &  39.81 / 17.19 / 26.57\\
        \textbf{\FWName} & 2.50 & \textbf{6.23} / \textbf{0.582} & \textbf{4.11} & \textbf{5.51} / \textbf{0.601} & \textbf{3.64} & \textbf{37.69 / 14.76 / 29.53} & \textbf{40.88 / 18.06 / 27.01} \\
        \midrule 
        QLoRA & 2.25 & 8.67 / 0.534 & 5.59 & 7.31 / 0.588 & 4.64 & 16.37 / 2.22 / 12.84 &  11.90 / 1.32 / 10.25 \\
        LoftQ & 2.25 & 8.22 / 0.543 & 5.24 & 6.96 / 0.564 & 4.46 & 32.71 / 10.94 / 25.37 & 39.36 / 16.87 / 26.29\\
        \textbf{\FWName} & 2.25 & \textbf{6.40} / \textbf{0.578}& \textbf{4.21} & \textbf{5.66} / \textbf{0.597} & \textbf{3.73} & \textbf{37.29 / 14.36 / 29.12} & \textbf{41.01 / 18.19 / 27.23} \\
        \midrule
        QLoRA & 2.00 & 9.17 / 0.526 & 6.07 & 7.64 / 0.551 & 5.02 & \textit{DNC} & 4.84 / 0.00 / 4.36 \\
        LoftQ & 2.00 & 8.63 / 0.536 & 5.68 & 7.27 / 0.558 & 4.75 & 31.89 / 10.18 / 24.59 & 38.88 / 16.49 / 25.85 \\
        \textbf{\FWName} & 2.00 & \textbf{6.60} / \textbf{0.574} & \textbf{4.35} & \textbf{5.79} / \textbf{0.593} & \textbf{3.84} & \textbf{36.75 / 13.93 / 28.61} & \textbf{40.15 / 17.48 / 26.67} \\
        \midrule
        QLoRA & 1.90 & -- & -- & -- & -- & -- & --  \\
        LoftQ & 1.90 & -- & -- & -- & -- & -- & --  \\
        \textbf{\FWName} & 1.90 & \textbf{7.13 / 0.562} & \textbf{4.94} & \textbf{6.16} / \textbf{0.583} & \textbf{4.22} & \textbf{34.05 / 11.74 / 26.49}  & \textbf{39.19 / 16.84 / 26.35}\\
        \midrule
        QLoRA & 1.80 & -- & -- & -- & -- & -- & -- \\
        LoftQ & 1.80 & -- & -- & -- & --  & -- & -- \\
        \textbf{\FWName} & 1.80 & \textbf{7.50 / 0.553} & \textbf{5.24} & \textbf{6.48 / 0.575} & \textbf{4.59} & \textbf{33.29 / 11.19 / 25.85} & \textbf{39.20 / 16.69 / 26.07}\\
        \midrule
        QLoRA & 1.75 & -- & -- & -- & -- & -- & -- \\
        LoftQ & 1.75 & -- & -- & -- & --  & -- & -- \\
        \textbf{\FWName} & 1.75 & \textbf{7.76 / 0.548} & \textbf{5.43} & \textbf{6.65} / \textbf{0.569} & \textbf{4.76} & \textbf{33.09 / 11.05 / 25.69} & \textbf{38.54 / 16.38 / 25.99}\\
        \bottomrule
    \end{tabular}
    } 
    \caption{\textbf{Performance comparison of different methods on LLaMA-2-7B, LLaMA-2-13B, and BART-large}. ``--" means this method fails to support this level of precision. ``\textit{DNC}" means fine-tuning fails to converge. \FWName (using PEFT) not only outperforms QLoRA and LoftQ in terms of performance-precision trade-off, but also enables us to fine-tune LLMs in the sub-2-bit range. Both QLoRA and LoftQ use NormalFloats \cite{dettmers2024qlora}. LoftQ results on Bart-Large are taken as the best of two strategies: (1) layers are ordered based sheerly on layer-index and (2) encoder layers are ordered before decoder layers. See Appendix \ref{app:loftqbart} for detailed results. Also, see Appendix \ref{app:ablation} for ablation analysis.}
    \vspace{-10pt}
    \label{tab:performance_comparison}
\end{table*}

\subsection{Evaluation Setup}

\paragraph{Hardware Platform} 
Experiments are conducted on NVIDIA A100 GPUs (80GB memory). 
Each LLaMA experiment runs on a single dedicated GPU.
Each BART-large experiment runs two instances concurrently on a single GPU.

\paragraph{Hyperparameters} 
For a fair comparison, we use identical hyperparameters across all methods, consistent with QLoRA~\cite{dettmers2024qlora} and LoftQ~\cite{li2023loftq}. Details on selected hyperparameters are in Appendix \ref{app:hyperparams}.

\paragraph{Language Models} 
We evaluate \FWName on a range of LLMs: LLaMA-2-7B, LLaMA-2-13B~\cite{touvron2023llama2}, BART-large~\cite{lewis2019bart}, and LLaMA-30B~\cite{touvron2023llama} (to assess ultra-low-bit scalability).

\paragraph{Datasets and Evaluation Metrics} 
We use standard datasets across different NLP tasks: WikiText-2~\cite{merity2016pointer} (language modeling, perplexity), Open-Assistant~\cite{kopf2024openassistant} (multi-turn conversation, perplexity), XSUM~\cite{narayan2018don} (summarization, ROUGE scores), and CNN/DailyMail~\cite{hermann2015teaching} (summarization, ROUGE scores).
Each dataset is evaluated using the standard metrics used in prior work.

\subsection{Baselines}

\paragraph{QLoRA} 
QLoRA originally employs a fixed 4-bit quantization for pretrained LLMs and does not support fine-tuning below 4 bits. 
To enable sub-4-bit QLoRA experiments, we follow the adaptation introduced in LoftQ. 
Additionally, QLoRA directly quantizes the pretrained weights while preserving their original distribution, initializing the low-rank tensors with zeros and small Gaussian noise.

\paragraph{LoftQ}
LoftQ performs mixed-precision quantization (2-bit/4-bit) and jointly optimizes both quantized LLM weights and low-rank adapter initialization. 
We match LoftQ's effective batch sizes but observe discrepancies with its published results due to:
\textbf{(1)} reproducibility constraints – the original authors did not release experiment seeds or hyperparameters used in mixed-precision trainings –
\textbf{(2)} hardware differences – Our experiments run on a single A100 GPU, whereas LoftQ was trained on 8 A100 GPUs with greater data parallelism - and 
\textbf{(3)} quantization implementation – The original LoftQ experiments rely on simulated quantization, which introduces discrepancies in quantized model weights, as noted by the research community\footnote{https://github.com/yxli2123/LoftQ/issues/7}. 
In contrast, we employ CUDA-kernel-based quantization and dequantization, ensuring more accurate and hardware-aligned results. 

\subsection{Analysis of Key Results}
\label{subsec:key_results}
Table \ref{tab:performance_comparison} presents the performance comparison of \FWName against QLoRA and LoftQ.
\begin{packeditemize}
    \item \textbf{Better performance at the same precision:} \FWName outperforms QLoRA and LoftQ across all sub-4-bit precision levels.
    In particular, at challenging 2-bit quantization, 
    \FWName achieves a perplexity reduction of: 2.21 (WikiText-2) / 1.45 (Open-Assistant) over QLoRA, and 1.76 (WikiText-2) / 1.12 (Open-Assistant) over LoftQ.
    \item \textbf{Same performance at lower precision:} 
    \FWName enables fine-tuning with 0.98 bits (QLoRA) / 0.76 bits (LoftQ) fewer per parameter (on average) without performance loss.
    For example, 2.5-bit \FWName on WikiText-2 (LLaMA-2-7B) matches 4-bit QLoRA; 1.9-bit \FWName on Open-Assistant (LLaMA-2-7B) matches 2.5-bit LoftQ. 
\end{packeditemize}

\subsection{Fine-Tuning LLMs with Ultra-Low Bits}
\label{subsec:ultra-low-bit}
We further explore: (1) Fine-tuning larger LLMs, and (2) fine-tuning under ultra-low-bit precision.
Results for LLaMA-30B at 1.15 and 1.25 bits are in Table \ref{tab:low_bit_experiment}.
\FWName is the first method to enable accurate LoRA fine-tuning at ultra-low-bit levels: 1.75 bits on LLaMA-2-7B\footnote{1.5-bit Llama-2-7B gives 9.38 perplexity on Wikitext2}, LLaMA-2-13B, and BART-large, or 1.15 bits on LLaMA-30B.

\subsection{Memory Implications}
\label{subsec:memory-imp}

Following the analysis methodology in QLoRA~\cite{dettmers2024qlora}, we evaluate the memory footprint of \FWName at different quantization precisions. Full visualizations are in Appendix \ref{app:memory}. Our results show that \FWName significantly reduces memory usage for both fine-tuning and inference, making ultra-low-bit LoRA practical on resource-constrained hardware.

For inference, reducing precision from 4-bit to 2-bit leads to 40\% lower memory usage on LLaMA-2-13B and 30\% on LLaMA-2-7B (Figures \ref{fig:inference_memory}). Compressing LLaMA-30B to 1.15 or 1.25 bits achieves even greater savings, reducing the memory footprint by 50\% (Figure \ref{fig:llama2_13b_memory}).

For fine-tuning, \FWName also achieves substantial reductions. Moving from 4-bit to 2-bit precision cuts memory consumption by 30\% on LLaMA-2-13B and 25\% on LLaMA-2-7B (Figures \ref{fig:finetuning_memory}). On LLaMA-30B, reducing precision to 1.15 or 1.25 bits per parameter leads to an estimated 45\% reduction in fine-tuning memory usage, making it feasible to train larger models under stricter memory constraints.

These memory savings are particularly impactful given that 4-bit QLoRA models are already highly compressed. By pushing below 2-bit precision with minimal performance loss, \FWName enables fine-tuning and deployment of LLMs on significantly smaller devices. For instance, a fine-tuned LLaMA-2-7B model can now be deployed on a Raspberry Pi 4 Model B (4GB RAM)~\cite{raspberry_pi_4_model_b}, making on-device inference feasible even in extreme resource-constrained settings. More strikingly, \FWName is the first method to enable LLaMA-30B fine-tuning on a single NVIDIA Tesla T4 (16GB VRAM)~\cite{nvidia_t4_virtualization_datasheet}, demonstrating its potential for democratizing large-scale LLM adaptation.

\begin{table}[ht]
    \centering
    \resizebox{0.5 \columnwidth}{!}{%
    \begin{tabular}{llccc}
        \toprule
        \multirow{2}{*}{\textbf{Bit}} & \multirow{2}{*}{\textbf{Dataset}} & \multicolumn{3}{c}{\textbf{Method}} \\
        \cmidrule(lr){3-5}
        & & QLoRA & LoftQ & \textbf{\FWName} \\
        \midrule
        \multirow{2}{*}{1.25} & WikiText-2       & --   & --   & 7.46 \\
         & Open-Assistant   & --   & --   & 5.44 \\
        \midrule
        \multirow{2}{*}{1.15} & WikiText-2       & --   & --   & 8.00 \\
         & Open-Assistant   & --   & --   & 5.73 \\
        \bottomrule
    \end{tabular}
    }
    \vspace{5pt}
    \caption{\
    \textbf{Performance of different methods on LLaMA-33B}. \FWName allows us to fine-tune LLMs at a precision level of as low as 1.15 bits, without significantly sacrificing accuracy.}
    \vspace{-15pt}
    \label{tab:low_bit_experiment}
\end{table}
\section{Conclusion and Future Work}

As LLMs scale, fine-tuning remains computationally and memory intensive, even with parameter-efficient methods like LoRA. 
We introduced \FWName{}, the first framework to enable accurate LoRA fine-tuning below 2 bits per parameter. 
By addressing key limitations in quantized LoRA, \FWName{} leverages fine-grained precision assignment, adaptive quantization mappings, and optimized CUDA kernels to minimize memory overhead while preserving performance.

Extensive evaluations show that \FWName{} achieves a superior performance–precision trade-off above 2 bits and remains accurate at even 1.15 bits per parameter, reducing memory usage by up to 50\%. 
This enables fine-tuning in ultra-resource-constrained environments, making LLMs more accessible for real-world applications.
Looking ahead, \FWName{} paves the way for ultra-low-bit fine-tuning and deployment. 
We hope it inspires further research and brings efficient LLM fine-tuning and inference to mobile devices, embedded systems, and beyond.


\bibliographystyle{plain}
\bibliography{main}

\appendix
\clearpage

\section{\FWName System Support for Low-Bit Fine-Grained LoRA Fine-tuning} \label{app:systemsupport}

In this appendix, we provide an overview of the CUDA-based system support we built for supporting low-bit fine-grained quantization and dequantization for LoRA fine-tuning. We specifically introduce the \textbf{\textit{Quantize}} and \textbf{\textit{Dequantize}} Kernels for low-bit fine-grained LoRA fine-tuning. We integrate this into the \textit{bitsandbytes}\footnote{https://github.com/bitsandbytes-foundation/bitsandbytes} library for usability.

\begin{figure}[ht]
\vspace{-14pt}
\begin{center}
\centerline{\includegraphics[width=0.9\columnwidth]{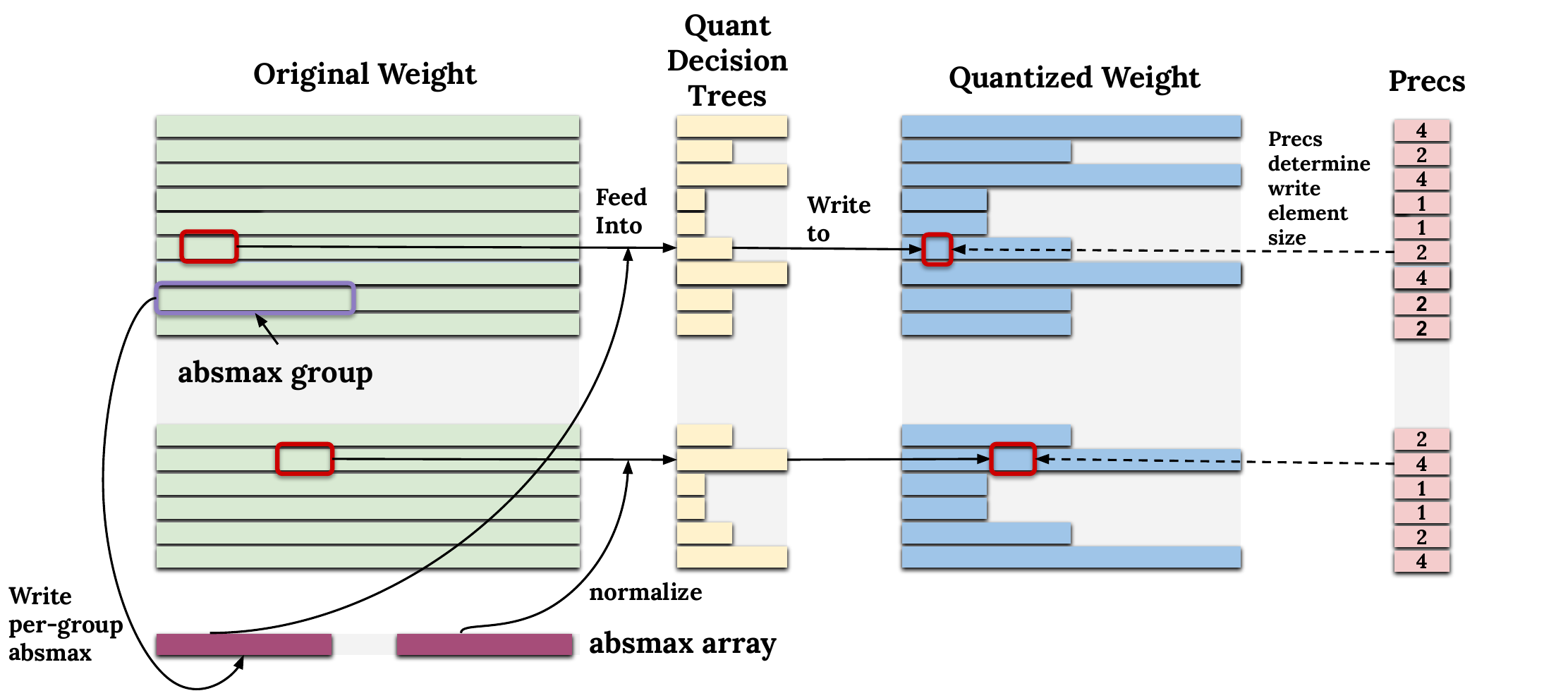}}
\vskip -0.2in
\caption{Overview of Kernel for Low-Bit Fine-Grained \textbf{\textit{Quantzation}}. Indexing logic omitted for simplicity.}
\label{fig:quant_kernel}
\end{center}
\vskip -0.4in
\end{figure}

\begin{figure}[ht]
\vspace{-14pt}
\begin{center}
\centerline{\includegraphics[width=0.85\columnwidth]{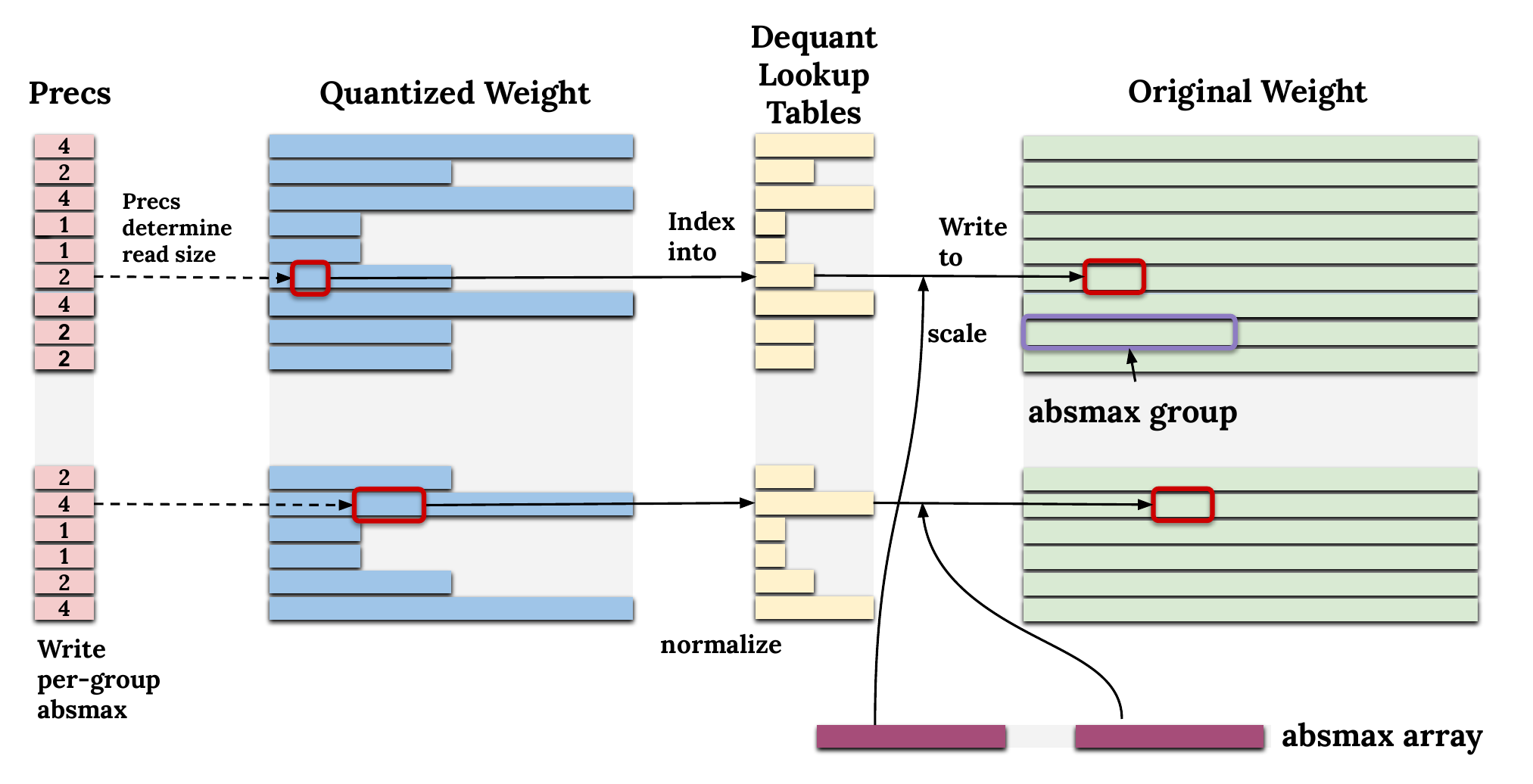}}
\vskip -0.2in
\caption{Overview of Kernel for Low-Bit Fine-Grained \textbf{\textit{Dequantization}}. Indexing logic omitted for simplicity.}
\label{fig:dequant_kernel}
\end{center}
\vskip -0.5in
\end{figure}

\subsection{Quantization (Figure \ref{fig:quant_kernel})}
During the quantization process, the kernel operates on each block of data from the weight tensor by loading it and computing its maximum magnitude, which is then stored and used for normalization. Next, each value in the block is quantized according to its channel’s bit precision (defined by \texttt{Precs}) and channel-specific decision boundaries (provided as arrays of decision trees). Finally, the packed, quantized results are written to the output buffer at the appropriate offset.

\subsection{Dequantization (Figure \ref{fig:dequant_kernel})}
During the dequantization process, each thread in the kernel calculates its offset into the quantized buffer based on the channel’s bit precision (\texttt{Precs}) and then loads the necessary packed bytes. Using the block’s \texttt{absmax} array, the thread rescales the dequantized values, which are obtained from a per-channel lookup table indexed by the unpacked quantized values. Depending on the precision (1, 2, or 4 bits), the kernel unpacks bits from the packed bytes, retrieves the corresponding float from the channel-specific lookup table, multiplies by \texttt{absmax}, and writes the final results into the output tensor.

\clearpage
\section{Ablation Studies} \label{app:ablation}
In this appendix, we perform ablation studies on the different components of \FWName{}. In particular, we compare the fine-tuning results with QLoRA, LoftQ, ours with only the precision assigner, and ours with both the precision assigner and the mapping/threshold searcher. Table \ref{tab:encdecablation} reports results of Bart-Large on XSUM and CNN/DailyMail. Table \ref{tab:llamaablation} reports results of Llama-2-7B on Wikitext2.

\begin{table}[ht!]
\centering
\resizebox{0.9\textwidth}{!}{%
\begin{tabular}{llcccccccccc}
\toprule
\multirow{2}{*}{\textbf{Technique}} & \multirow{2}{*}{\textbf{Metric}} 
& \multicolumn{5}{c}{\textbf{XSUM}} 
& \multicolumn{5}{c}{\textbf{CNN/DailyMail}} \\
\cmidrule(lr){3-7}\cmidrule(lr){8-12}
 & & \textbf{2} & \textbf{2.25} & \textbf{2.5} & \textbf{3} & \textbf{4}
   & \textbf{2} & \textbf{2.25} & \textbf{2.5} & \textbf{3} & \textbf{4} \\
\midrule

\multirow{3}{*}{\textbf{QLoRA}} 
 & \textbf{ROUGE1$\uparrow$} 
   & DNC & 16.3727 & 15.3282 & 17.6011 & 39.0742 
   & 4.8420 & 11.8987 & 13.6767 & 15.3374 & 41.1846 \\
 & \textbf{ROUGE2$\uparrow$} 
   & DNC & 2.2212 & 1.9712 & 2.6801 & 16.3124 
   & 0.0040 & 1.3188 & 1.0364 & 1.1239 & 18.3249 \\
 & \textbf{ROUGEL$\uparrow$} 
   & DNC & 12.8367 & 12.5547 & 13.9294 & 31.0891 
   & 4.3608 & 10.2535 & 9.9929 & 10.4375 & 27.5773 \\
\midrule

\multirow{3}{*}{\textbf{LoftQ}} 
 & \textbf{ROUGE1$\uparrow$} 
   & 31.8941 & 32.6153 & 33.7645 & 36.0222 & 40.3429
   & 38.8866 & 39.3648 & 39.8138 & 40.4684 & 41.1247 \\
 & \textbf{ROUGE2$\uparrow$} 
   & 10.1775 & 10.7005 & 11.7335 & 13.4883 & 17.0615
   & 16.4935 & 16.8711 & 17.1934 & 17.7529 & 18.2853 \\
 & \textbf{ROUGEL$\uparrow$} 
   & 24.5908 & 25.1533 & 26.2388 & 28.1558 & 31.9186
   & 25.8534 & 26.2877 & 26.5726 & 26.8812 & 27.5372 \\
\midrule

\multirow{3}{*}{\textbf{Ours, PA Only}} 
 & \textbf{ROUGE1$\uparrow$} 
   & 31.8941 & 36.4856 & 37.2861 & 38.1543 & 40.3429 
   & 38.8866 & 40.3669 & 40.6187 & 41.1820 & 41.1247 \\
 & \textbf{ROUGE2$\uparrow$} 
   & 10.1775 & 13.6310 & 14.3775 & 15.1856 & 17.0615 
   & 16.4935 & 17.6272 & 17.8600 & 18.3966 & 18.2853 \\
 & \textbf{ROUGEL$\uparrow$} 
   & 24.5908 & 28.4593 & 29.2036 & 29.9965 & 31.9186 
   & 25.8534 & 26.7442 & 26.9107 & 27.4180 & 27.5372 \\
\midrule

\multirow{3}{*}{\textbf{Ours, PA + MTSearch}} 
 & \textbf{ROUGE1$\uparrow$} 
   & 36.7454 & 37.2915 & 37.6897 & 38.8396 & 40.2669
   & 40.1489 & 41.0133 & 40.8819 & 40.8470 & 40.9493 \\
 & \textbf{ROUGE2$\uparrow$} 
   & 13.9324 & 14.3641 & 14.7587 & 15.6822 & 17.1800
   & 17.4843 & 18.1898 & 18.0609 & 18.1191 & 18.1223 \\
 & \textbf{ROUGEL$\uparrow$} 
   & 28.6133 & 29.1175 & 29.5275 & 30.5712 & 32.0614
   & 26.6717 & 27.2268 & 27.0113 & 27.2309 & 27.5392 \\
\bottomrule
\end{tabular}
}

\caption{Comparison of ROUGE scores with \textbf{Bart-Large} on \textbf{XSUM} and \textbf{CNN/DailyMail }under different techniques and combinations of techniques. \textbf{PA} refers to the two-level precision assigner. \textbf{MTSearch} refers to mapping and threshold search.}
\label{tab:encdecablation}
\vskip -0.2in
\end{table}

\begin{table}[ht!]
\centering
\resizebox{0.75\textwidth}{!}{%
\begin{tabular}{l l c c c c c c c c c}
\toprule

\multirow{2}{*}{\textbf{Technique}} & \multirow{2}{*}{\textbf{Metric}} 
& \multicolumn{9}{c}{\textbf{Wikitext2}} \\
\cmidrule(lr){3-11}
 & & \textbf{1.5} & \textbf{1.75} & \textbf{1.8} & \textbf{1.9} 
   & \textbf{2} & \textbf{2.25} & \textbf{2.5} & \textbf{3} & \textbf{4} \\
\midrule

\multirow{2}{*}{\textbf{QLoRA}} 
 & Perplexity & - & - & - & - & 9.17 & 8.67 & 8.05 & 7.13 & 6.22 \\
 & Accuracy   & - & - & - & - & 0.526 & 0.534 & 0.546 & 0.566 & 0.583 \\
\midrule

\multirow{2}{*}{\textbf{LoftQ}} 
 & Perlexity & - & - & - & - & 8.63 & 8.22 & 7.72 & 6.87 & 5.26 \\
 & Accuracy  & - & - & - & - & 0.536 & 0.543 & 0.552 & 0.571 & 0.613 \\
\midrule

\multirow{2}{*}{\textbf{PA Only}} 
 & Perplexity & ND & ND & ND & ND & 8.63 & 8.22 & 7.75 & 6.75 & 5.26 \\
 & Accuracy   & ND & ND & ND & ND & 0.536 & 0.542 & 0.550 & 0.570 & 0.613 \\
\midrule

\multirow{2}{*}{\shortstack[l]{\textbf{PA}+\textbf{MTSearch}}} 
 & Perplexity & 9.38 & 7.76 & 7.50 & 7.13 & 6.60 & 6.40 & 6.23 & 5.84 & 5.25 \\
 & Accuracy   & 0.521 & 0.548 & 0.553 & 0.562 & 0.574 & 0.578 & 0.582 & 0.593 & 0.6123 \\
\bottomrule
\end{tabular}
} 
\caption{Comparison of perplexity and accuracy of \textbf{Llama2-7B} on the \textbf{Wikitext2} dataset under different techniques and combinations of techniques. ``ND'' indicates no data, ``-'' indicates the techniques fail to support these precisions. \textbf{PA} refers to the two-level precision assigner. \textbf{MTSearch} refers to mapping and threshold search.}
\label{tab:llamaablation}
\end{table}

We observe generally that the precision assigner gives a significant advantage for Bart-Large on summarization summarization tasks (Table \ref{tab:encdecablation}), whereas the mapping/threshold searcher yields significant gains for Llama-2-7B on Wikitext2 (Table \ref{tab:llamaablation}). Interestingly, the precision assigner only yields minimal advantage over LoftQ for Llama-2-7B on Wikitext2. 

In terms of applying the mapping/threshold searcher to Bart-Large (Table \ref{tab:encdecablation}), we see a less significant gain in the 2.5-4.0 precision range in comparison to the gains in the 2.0-2.5 precision range. This is in line with our intuition as at higher precision there are more mappings and thresholds for capturing distributions during quantization and dequantization; thereby, the tolerance for a less accurate combination of mappings and thresholds is higher.

We encourage future works to build upon \FWName{} and study the optimal precision assignment and mapping/threshold search algorithms for different types of architectures (\ie encoder-decoder, encoder-only, decoder-only). We will provide easy integration to new advances in our open-sourced repository.
\clearpage
\section{Memory Requirements}\label{app:memory}
\begin{figure}[ht]
\vskip 0.2in
\begin{center}
\centerline{\includegraphics[width=0.9\textwidth]{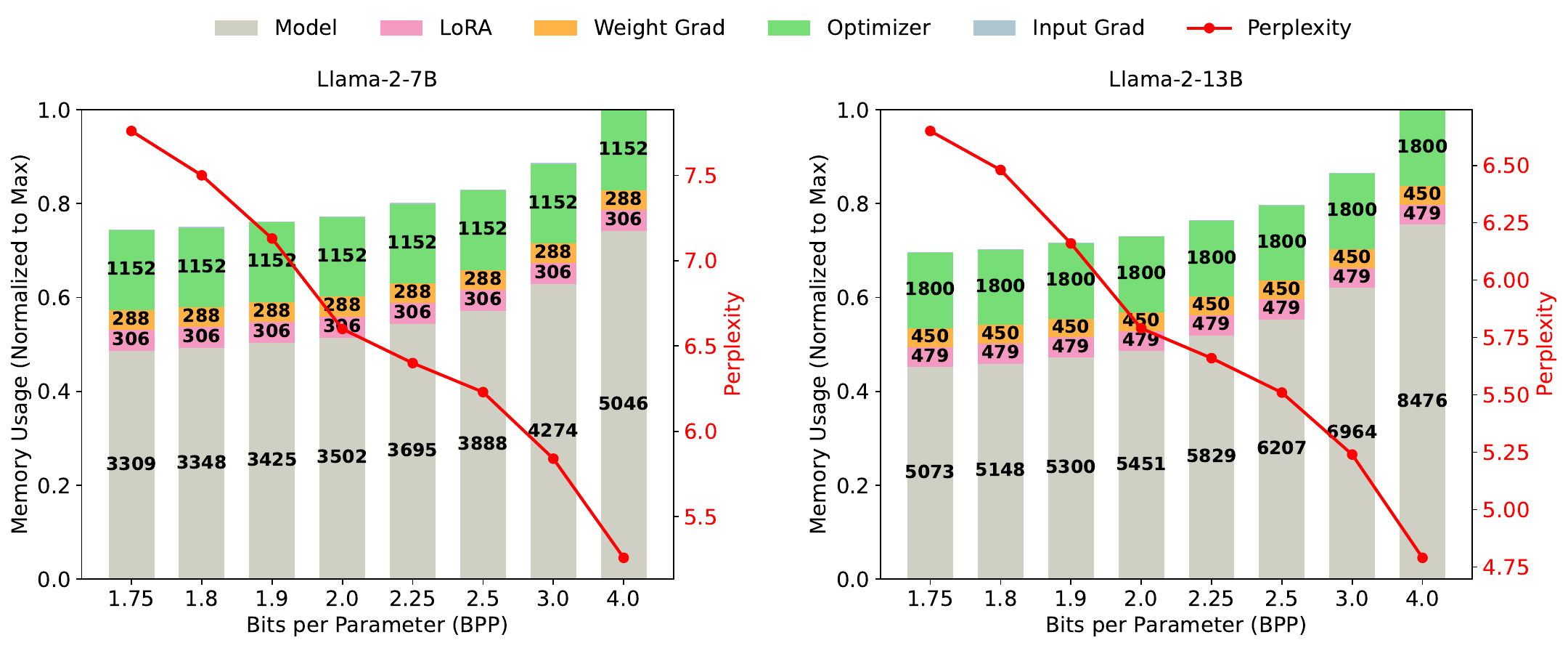}}
\caption{Decomposition of \textbf{\textit{finetuning}} memory footprint for Llama-2 7B and 13B under different bits per parameter.}
\label{fig:finetuning_memory}
\end{center}
\vskip -0.2in
\end{figure}

\begin{figure}[ht]
\vskip 0.2in
\begin{center}
\centerline{\includegraphics[width=0.9\textwidth]{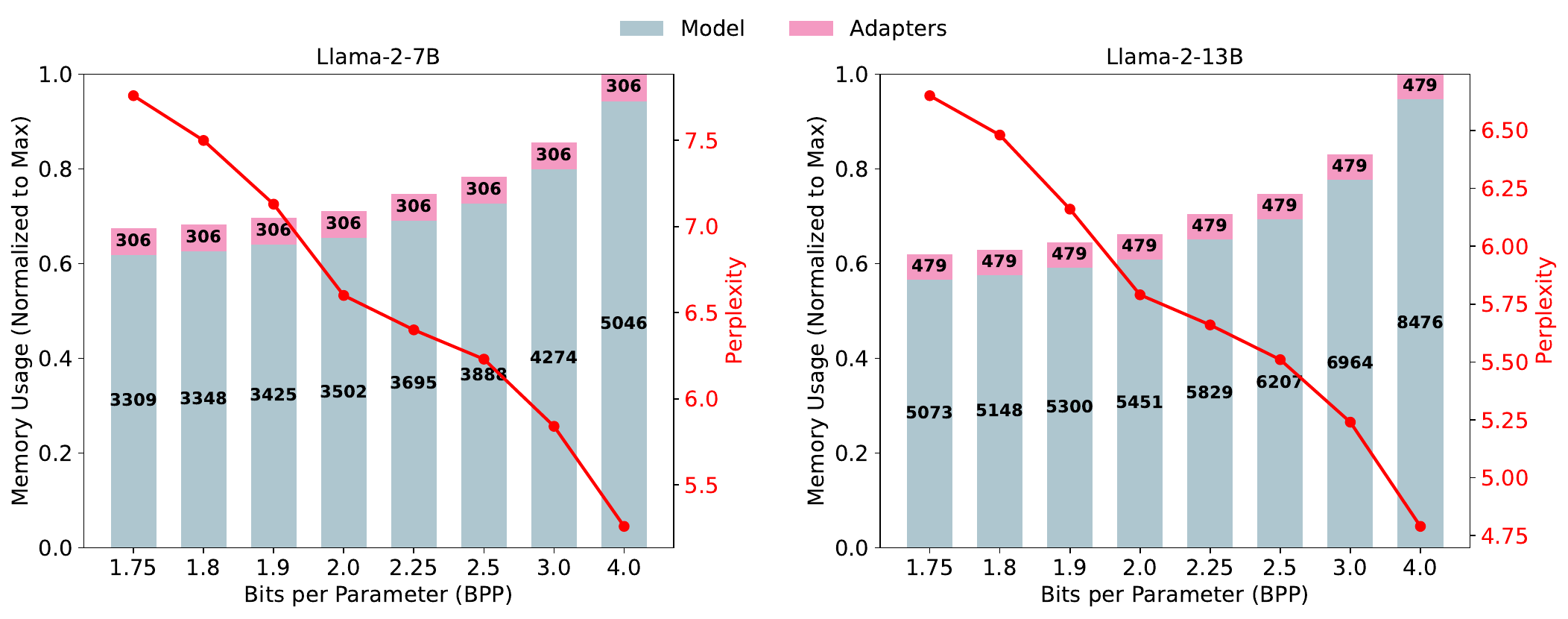}}
\caption{Decomposition of \textbf{\textit{inference}} memory footprint for Llama-2 7B and 13B under different bits per parameter.}
\label{fig:inference_memory}
\end{center}
\vskip -0.2in
\end{figure}

\begin{figure}[ht]
\vskip 0.2in
\begin{center}
\centerline{\includegraphics[width=0.85\textwidth]{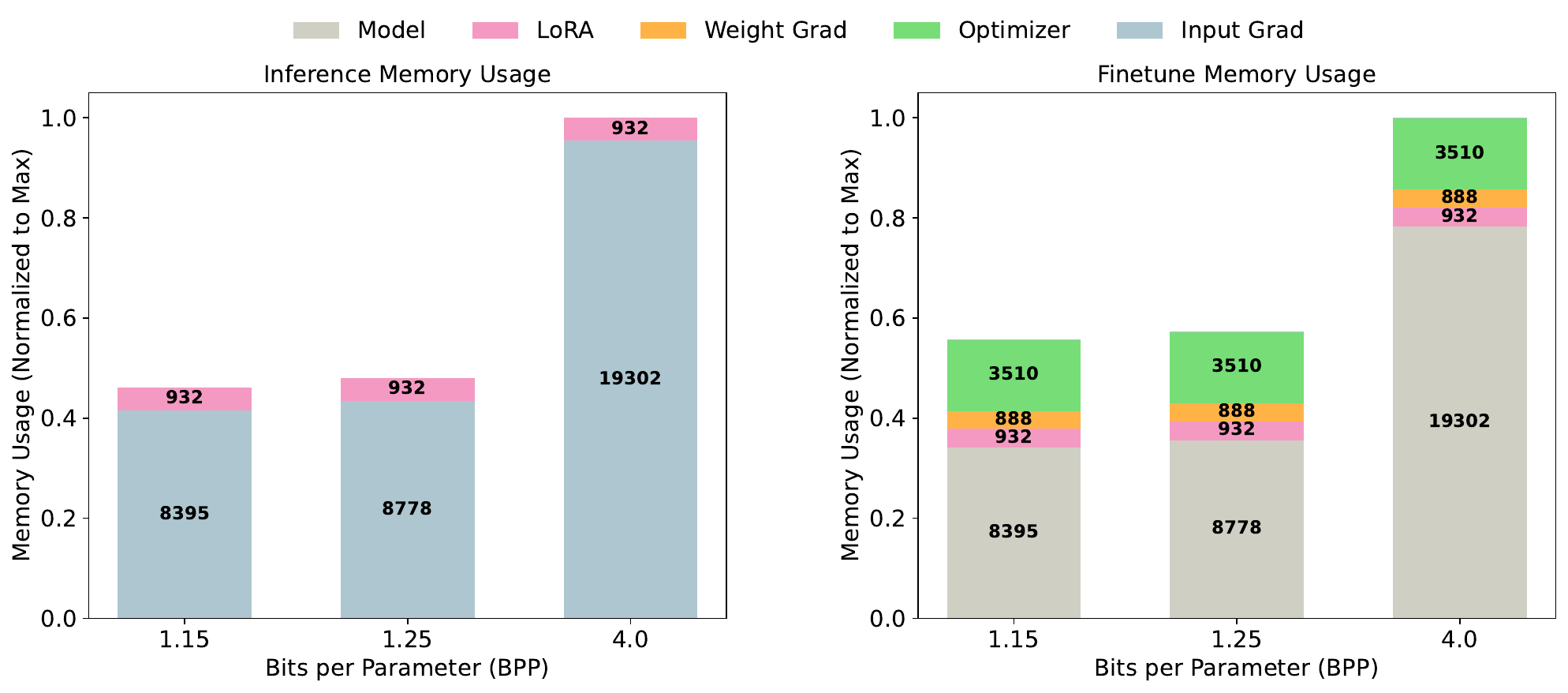}}
\caption{Decomposition of memory footprint for Llama-33B under different bits per parameter.}
\label{fig:llama2_13b_memory}
\end{center}
\vskip -0.2in
\end{figure}

In this appendix, we provide a detailed breakdown of memory footprints. 
Figure \ref{fig:finetuning_memory} shows the \textbf{\textit{fine-tuning}} memory footprint decompositions for \textbf{Llama-2-7B} and \textbf{Llama-2-13B}. Figure \ref{fig:inference_memory} shows the \textbf{\textit{inference}} memory decomposition for \textbf{Llama-2-7B} and \textbf{Llama-2-13B}. Figure \ref{fig:llama2_13b_memory} shows \textbf{\textit{both inference and fine-tuning}} memory footprint decomposition for \textbf{Llama-33B}.  For finetuning, we follow QLoRA's setup of using a batch size of 1 and sequence length of 512. Number labels on the bars are in MegaBytes (MB). Estimations are linear layer only (not attention).

\newpage
\clearpage

\section{Low-Rank Initializers: LoftQ vs PiSSA} \label{app:lowrankinit}
Both LoftQ~\cite{li2023loftq} and PiSSa~\cite{meng2024pissa} use an iterative two-step process to enhance LoRA finetuning by exploiting low-rank structure and quantizing the residual. In each iteration:

\begin{enumerate}
    \item \textbf{Low-Rank Decomposition:} A singular value decomposition (SVD) of the current weight (or an updated version of it) is performed to factor out a low-rank approximation.
    \item \textbf{Residual Quantization:} The remaining component (i.e., the difference between the original weight and the low-rank approximation) is quantized to preserve overall model capacity with fewer bits.
\end{enumerate}

The key distinction lies in how these two steps are ordered:
\begin{itemize}
    \item \textbf{LoftQ} first quantizes the residual, which at the beginning is simply the full base weight. Thus, it ``initializes'' by treating the entire unmodified weight as a residual to be quantized and only then proceeds with the low-rank factorization in subsequent iterations.
    \item \textbf{PiSSA} starts by performing SVD on the unquantized base weight, extracting a low-rank representation before any quantization. Only after factoring out the low-rank component does PiSSa quantize the remaining residual. 
\end{itemize}

Below (Table \ref{tab:full_init_compare}) are the experimental results covering the full 2.0-to-4.0 range comparing these two initialization techniques. From our replication, LoftQ consistently outperforms PiSSa in terms of final task performance. We adopt LoftQ's mixed precision scheme as a result.

\begin{table}[ht]
    \centering
    \scalebox{0.8}{%
    \begin{tabular}{llcccccccccc}
        \toprule
        \multicolumn{2}{l}{\textbf{Setup}} & \multicolumn{5}{c}{\textbf{Llama-7b}} & \multicolumn{5}{c}{\textbf{Llama-13b}}\\
        \cmidrule(lr){1-2} \cmidrule(lr){3-7} \cmidrule(lr){8-12}
        \textbf{Method} & \textbf{Dataset} & \textbf{2.0} & \textbf{2.25} & \textbf{2.5} & \textbf{3.0} & \textbf{4.0} & \textbf{2.0} & \textbf{2.25} & \textbf{2.5} & \textbf{3.0} & \textbf{4.0} \\
        \midrule
        PiSSA & \multirow{2}{*}{WikiText-2 ($\downarrow$)} & 1919.63 & 795.88& 1938.33 &1397.26 &  5.53 & 1825.68 & 1822.62 & 1769.34 & 1549.48 & 5.05 \\
        LoftQ &  & 8.63 & 8.22 & 7.72  & 6.87 & 5.26 & 7.27 & 6.96 & 5.7 & 5.91 & 4.79 \\
        \bottomrule
    \end{tabular}%
    }
    \caption{Perplexities of PiSSA and LoftQ as initialization methods on WikiText-2 covering full range from 2.0 to 4.0. Lower values indicate better performance ($\downarrow$).}
    \label{tab:full_init_compare}
\end{table}

\clearpage
\section{Overheads of Mapping/Thresholds Searcher and Precision Assigner}\label{app:overhead}

In this appendix, we report the overhead incurred by \FWName{'s} newly added components. Note that \FWName{'s} mapping/threshold learner and precision learner can both be done offline. The former only needs to be run for each model architecture, while the latter only needs to be run for each combination of model architecture and user-specified precision requirement.

\textbf{Mapping/Threshold Learner Overhead} We timed the overhead of our mapping/threshold learner on a single NVIDIA A100 SXM GPU with GPU memory. For each precision, we ran 2 iterations of the Lloyd-Max algorithm, which we found sufficient to push down the MSE and enhance task performance.  From the average of 10 runs, each run on Bart-Large, Llama2-7B, and Llama2-13B takes 111.46, 309.87, and 464.92 seconds, respectively.

\textbf{Solver Overhead} We build our two-level ILP pipeline using the opensourced Coin-Or Branch and Cut (CBC) \cite{saltzman2002coin} solver via the Python-based modeling library PuLP \cite{mitchell2011pulp}. The experiments were run on a server with 2x Intel Xeon Gold 6342 CPUs. The solver uses 8 threads in parallel. We timed the overhead of running this two-level ILP pipeline for each combination of model architecture and precision requirement. From the average of 10 runs, each run on Bart-Large, Llama2-7B, and Llama2-13B takes 48.99 seconds, 319.13 seconds, and 665.93 seconds, respectively. Note that an exact (i.e., one-step) solver for the same problem would fail to finish in a reasonable amount of time. 

\clearpage
\section{Initialization of the Mapping and Threshold Learner} \label{app:lloydinit}

We attach the mappings and thresholds we use for the weighted Lloyd-Max algorithm in Listing \ref{lst:ref-init}.

\begin{listing}[ht]
\begin{lstlisting}[
    language=Python,
    basicstyle=\ttfamily\small,
    numberstyle=\tiny\color{gray},
    frame=single,
    backgroundcolor=\color{gray!15},
    captionpos=b
]
mappings_4bit_init = torch.tensor(
    [[-1.0, -0.6961928, -0.5250731, -0.3949175, -0.28444138,
      -0.18477343, -0.09105, 0.0, 0.0795803, 0.1609302,
      0.2461123, 0.33791524, 0.44070983, 0.562617, 0.72295684, 1.0]],
    dtype=torch.float32, 
    device=device
).repeat(nchannels, 1)

thresholds_2bit_init = torch.tensor(
    [[-0.5, 0.16895762, 0.66895762]],
    dtype=torch.float32, 
    device=device
).repeat(nchannels, 1)

mappings_2bit_init = torch.tensor(
    [[-1.0, 0.0, 0.3379, 1.0]],
    dtype=torch.float32, 
    device=device
).repeat(nchannels, 1)

thresholds_1bit_init = torch.tensor(
    [[0.0]],
    dtype=torch.float32, 
    device=device
).repeat(nchannels, 1)

mappings_1bit_init = torch.tensor(
    [[-1.0, 1.0]],
    dtype=torch.float32, 
    device=device
).repeat(nchannels, 1)
\end{lstlisting}
\caption{Initializations for bit-precision mappings and thresholds.}
\label{lst:ref-init}
\end{listing}

\newpage
\clearpage
\section{Weighted Lloyd-Max Algorithm}
\label{sec:lloyd-max}

In this subsection, we briefly introduce the Weighted Lloyd-Max Algorithm, extended from the original Lloyd-Max algorithm \cite{lloyd1982least,max1960quantizing}.

Let \(\{x_i\}_{i=1}^N \subset \mathbb{R}^d\) be data points with corresponding weights \(\{w_i\}_{i=1}^N\), \(w_i > 0\). We seek to find \(K\) cluster centers \(\{y_j\}_{j=1}^K \subset \mathbb{R}^d\) minimizing the weighted mean-squared error:
\[
\min_{\{c(i)\}, \{y_j\}} 
\sum_{i=1}^N w_i \,\bigl\|x_i - y_{c(i)}\bigr\|^2,
\]
where \(c(i) \in \{1, \dots, K\}\) is the cluster index assigned to \(x_i\). The weighted Lloyd's algorithm alternates between:

\noindent\textbf{1.\ Assignment (E-step):} Assign each data point \(x_i\) to the cluster center closest in Euclidean distance:
\begin{equation}
  c(i) \;\leftarrow\; \underset{1 \le j \le K}{\mathrm{arg\,min}}\; \bigl\| x_i - y_j \bigr\|^2.
  \tag{E-step}\label{eq:e-step}
\end{equation}

\noindent\textbf{2.\ Update (M-step):} Recompute each cluster center \(y_j\) as the weighted centroid of the points assigned to it:
\begin{equation}
  y_j \;\leftarrow\; \frac{\sum_{i : c(i) = j} w_i \, x_i}{\sum_{i : c(i) = j} w_i}.
  \tag{M-step}\label{eq:m-step}
\end{equation}

These steps are repeated until convergence or until a stopping criterion (e.g., a maximum number of iterations) is met.
\section{Parameter Variances Along Output Channel Dimension \textit{vs} Along Input Channel Dimension \label{app:channel_var}}
\begin{figure}[ht]
\vskip 0.2in
\begin{center}
\centerline{\includegraphics[width=0.5\columnwidth]{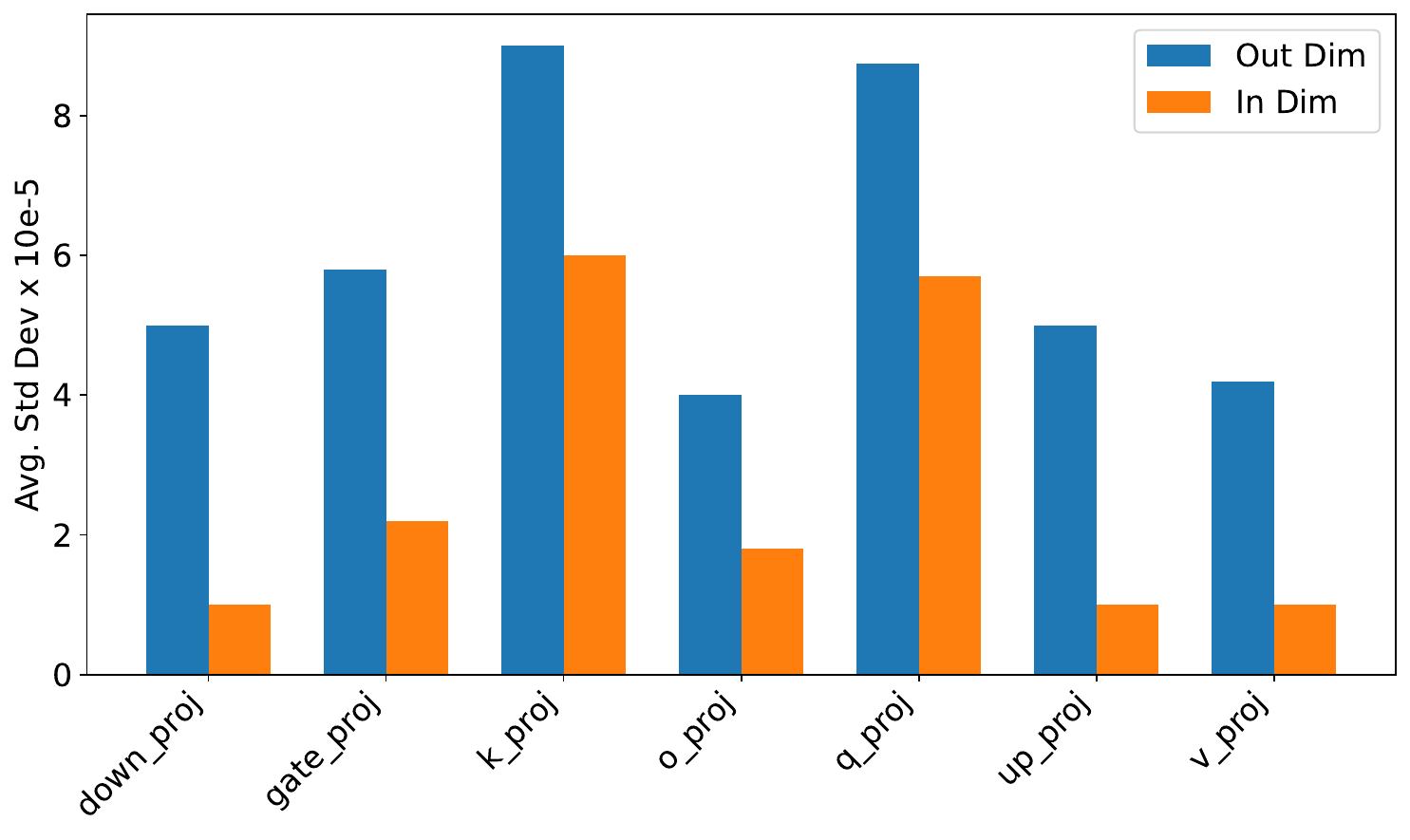}}
\caption{Standard Deviations Along Output Channel Dimension vs Along Input Channel Dimension, Grouped by Layer Type.}
\label{fig:channel_var}
\end{center}
\vskip -0.2in
\end{figure}

Figure~\ref{fig:channel_var} compares weight variance patterns across different layer types by examining the standard deviation along both input and output channel dimensions. We observe that the standard deviation is consistently higher along the output channel dimension than the input channel dimension.

\clearpage
\clearpage

\section{Experiment Hyperparamter Setup}\label{app:hyperparams}

In this appendix, we lay out the hyperparameters used for different experiments. Table~\ref{tab:wikitext2_hparams} details the settings for Llama on the Wikitext-2 dataset. Table~\ref{tab:cnn_dailymail_hparams} provides the configuration used for fine-tuning Bart-Large on CNN/DailyMail, and Table~\ref{tab:xsum_hparams} does the same for XSUM. Finally, Table~\ref{tab:qlora_hparams} describes the hyperparameters for Llama on the Open Assistant (oasst1) dataset.
\begin{table}[ht]
\centering
\begin{tabular}{ll}
\toprule
\textbf{Hyperparameter} & \textbf{Value} \\
\midrule
\texttt{model\_name\_or\_path} & meta-llama/Llama-2-7b-hf \\
\texttt{data\_seed}            & 42 \\
\texttt{evaluation\_strategy}  & steps \\
\texttt{eval\_dataset\_size}   & 1024 \\
\texttt{max\_eval\_samples}    & 1000 \\
\texttt{per\_device\_eval\_batch\_size} & 4 \\
\texttt{dataloader\_num\_workers} & 3 \\
\texttt{lora\_r}               & 64 \\
\texttt{lora\_alpha}           & 64 \\
\texttt{lora\_modules}         & all \\
\texttt{bf16}                  & True \\
\texttt{warmup\_ratio}         & 0.03 \\
\texttt{lr\_scheduler\_type}   & cosine \\
\texttt{gradient\_checkpointing} & True \\
\texttt{dataset}               & wikitext \\
\texttt{dataset\_config}       & wikitext-2-raw-v1 \\
\texttt{per\_device\_train\_batch\_size} & 16 \\
\texttt{gradient\_accumulation\_steps}   & 4 \\
\texttt{max\_steps}            & 126 \\
\texttt{eval\_steps}           & 20 \\
\texttt{learning\_rate}        & 0.0003 \\
\texttt{adam\_beta2}           & 0.999 \\
\texttt{max\_grad\_norm}       & 0.3 \\
\texttt{weight\_decay}         & 0.1 \\
\texttt{seed}                  & 0 \\

\texttt{block\_size}           & 1024 \\
\bottomrule
\end{tabular}
\caption{Hyperparameters used for all Llama experiments on Wikitext-2.}
\label{tab:wikitext2_hparams}
\end{table}

\begin{table}[ht!]
\centering
\begin{tabular}{ll}
\toprule
\textbf{Hyperparameter} & \textbf{Value} \\
\midrule
\texttt{learning\_rate} & 1e-4 \\
\texttt{seed} & 11 \\
\texttt{dataset\_name} & cnn\_dailymail \\
\texttt{dataset\_config} & ``3.0.0" \\
\texttt{pad\_to\_max\_length} & True \\
\texttt{max\_source\_length} & 512 \\
\texttt{num\_train\_epochs} & 15 \\
\texttt{per\_device\_train\_batch\_size} & 8 \\
\texttt{per\_device\_eval\_batch\_size} & 32 \\
\texttt{gradient\_accumulation\_steps} & 32 \\
\texttt{model\_name\_or\_path} & facebook/bart-large \\
\texttt{evaluation\_strategy} & epoch \\
\texttt{predict\_with\_generate} & True \\
\bottomrule
\end{tabular}
\caption{Hyperparameters for fine-tuning Bart-Large on CNN/DailyMail.}
\label{tab:cnn_dailymail_hparams}
\end{table}

\begin{table}[ht!]
\centering
\begin{tabular}{ll}
\toprule
\textbf{Hyperparameter} & \textbf{Value} \\
\midrule
\texttt{learning\_rate} & 1e-4 \\
\texttt{seed} & 11 \\
\texttt{dataset\_name} & xsum \\
\texttt{dataset\_config} & ``3.0.0" \\
\texttt{pad\_to\_max\_length} & True \\
\texttt{max\_source\_length} & 512 \\
\texttt{num\_train\_epochs} & 25 \\
\texttt{per\_device\_train\_batch\_size} & 4 \\
\texttt{per\_device\_eval\_batch\_size} & 32 \\
\texttt{gradient\_accumulation\_steps} & 32 \\
\texttt{model\_name\_or\_path} & facebook/bart-large \\
\texttt{evaluation\_strategy} & epoch \\
\bottomrule
\end{tabular}
\caption{Hyperparameters for fine-tuning Bart-Large on XSUM.}
\label{tab:xsum_hparams}
\end{table}

\begin{table}[ht!]
\centering
\begin{tabular}{ll}
\toprule
\textbf{Hyperparameter} & \textbf{Value} \\
\midrule
\texttt{model\_name\_or\_path}      & meta-llama/Llama-2-7b-hf \\
\texttt{data\_seed}                 & 42 \\
\texttt{evaluation\_strategy}       & steps \\
\texttt{eval\_dataset\_size}        & 1024 \\
\texttt{max\_eval\_samples}         & 500 \\
\texttt{per\_device\_eval\_batch\_size} & 1 \\
\texttt{max\_new\_tokens}           & 32 \\
\texttt{dataloader\_num\_workers}   & 3 \\
\texttt{group\_by\_length}          & True \\
\texttt{logging\_strategy}          & steps \\
\texttt{remove\_unused\_columns}    & False \\
\texttt{lora\_r}                    & 64 \\
\texttt{lora\_alpha}                & 64 \\
\texttt{lora\_modules}              & all \\
\texttt{bf16}                       & True \\
\texttt{warmup\_ratio}              & 0.03 \\
\texttt{lr\_scheduler\_type}        & constant \\
\texttt{gradient\_checkpointing}    & True \\
\texttt{dataset}                    & oasst1 \\
\texttt{source\_max\_len}           & 16 \\
\texttt{target\_max\_len}           & 512 \\
\texttt{per\_device\_train\_batch\_size} & 4 \\
\texttt{gradient\_accumulation\_steps}   & 4 \\
\texttt{max\_steps}                 & 1875 \\
\texttt{eval\_steps}                & 200 \\
\texttt{learning\_rate}             & 0.0002 \\
\texttt{adam\_beta2}                & 0.999 \\
\texttt{max\_grad\_norm}            & 0.3 \\
\texttt{lora\_dropout}              & 0.1 \\
\texttt{weight\_decay}              & 0.0 \\
\texttt{seed}                       & 0 \\
\bottomrule
\end{tabular}
\caption{Hyperparameters used for all Llama experiments on Open Assistant (oasst1).}
\label{tab:qlora_hparams}
\end{table}

\newpage
\clearpage
\section{Github Issues Related To The Lack of A Practical Quantization Primitive}\label{app:githubissues}

\begin{itemize}
    \item \url{https://github.com/yxli2123/LoftQ/issues/1}: The use of NF fake quantization in LoftQ
    \item \url{https://github.com/yxli2123/LoftQ/issues/7}: Discrepancy between real model weights and expected model weights due to fake quantization in LoftQ
    \item \url{https://github.com/yxli2123/LoftQ/issues/23}: The use of NF fake quantization in LoftQ
    \item \url{https://github.com/yxli2123/LoftQ/issues/36}: Unrealized GPU memory saving due to fake quantization in LoftQ
    \item \url{https://github.com/GraphPKU/PiSSA/issues/30}: Unrealized LLM model size reduction due to fake quantization in PiSSA
\end{itemize}
\section{Detailed LoftQ Results on Bart-Large}\label{app:loftqbart}

In this appendix, Table~\ref{tab:loftq_2version_results} presents results for LoftQ-based fine-tuning under two interpretations: (1) ordering layers by index (\emph{Layer Index}), and (2) treating encoder layers as preceding decoder layers (\emph{Encoder First}). For fairness, we select the best-performing interpretation as the LoftQ baseline. Note that for average precision = 2.0 and 4.0, both interpretations lead to the same implementation. Interestingly, \emph{Encoder First} consistently performs better on XSUM, whereas \emph{Layer Index} generally achieves higher scores on CNN/DailyMail.

\begin{table}[ht]
    \centering
    \scalebox{0.6}{%
    \begin{tabular}{llccccc}
        \toprule
        \multicolumn{2}{l}{\textbf{Setup}} & \multicolumn{5}{c}{\textbf{Bart-Large}} \\
        \cmidrule(lr){1-2} \cmidrule(lr){3-7}
        \textbf{Method} & \textbf{Dataset} & \textbf{2.0} & \textbf{2.25} & \textbf{2.5} & \textbf{3.0} & \textbf{4.0} \\
        \midrule
        \textit{Layer Index} & \multirow{2}{*}{XSUM($\uparrow$)} & \multirow{2}{*}{\textbf{31.89/10.18/24.59}} & 32.62/10.70/25.15 & 33.76/11.73/26.24 & 36.02/13.49/28.16 & \multirow{2}{*}{\textbf{40.34/17.06/31.92}} \\
        \textit{Encoder First} & & & \textbf{32.72/10.94/25.38} & \textbf{34.49/12.26/27.05} & \textbf{37.23/14.34/29.31} & \\
        \midrule
        \textit{Layer Index} & \multirow{2}{*}{CNN/DailyMail($\uparrow$)} & \multirow{2}{*}{\textbf{38.89/16.49/25.85}} & \textbf{39.36/16.87/26.29} & \textbf{39.81/17.19/26.57} & \textbf{40.47/17.75/26.88} & \multirow{2}{*}{\textbf{41.12/18.29/27.54}} \\
        \textit{Encoder First} & & & 39.27/16.84/26.26 & 39.53/17.05/26.45 & 39.89/17.36/26.73 & \\
        \bottomrule
    \end{tabular}%
    }
    \caption{ROUGE-1/ROUGE-2/ROUGE-L results of Bart-Large fine-tuned with LoftQ. Columns labeled 2.0, 2.25, 2.5, 3.0, and 4.0 refer to the average bits per parameter (bpp) or average precision. Higher scores indicate better performance.}
    \label{tab:loftq_2version_results}
\end{table}

\end{document}